\documentclass[10pt,twocolumn,letterpaper]{article}

\pdfoutput=1

\usepackage[pagenumbers]{cvpr} % To force page numbers, e.g. for an arXiv version

\usepackage[dvipsnames,table]{xcolor}
\usepackage{graphicx}
\usepackage{amsmath}
\usepackage{amssymb}
\usepackage{booktabs}
\usepackage{float}
\usepackage{enumerate}
\usepackage[all]{nowidow}
\usepackage{uoftcolors}

\usepackage{mathtools} % amsmath with fixes and additions
\usepackage{booktabs} % commands to create good-looking tables
\usepackage{tikz} % nice language for creating drawings and diagrams
\usepackage{amsfonts}
\usepackage{multirow}
\usepackage{multicol}
\usepackage{color,colortbl}
\usepackage{xspace}
\usepackage{wrapfig}
\usepackage{subcaption}
\usepackage{bbm}
\usepackage{bm}
\usepackage{url}

\def\eg{e.g.,\ }               % for example
\def\ie{i.e.,\ }               % that is, in other words
\def\vs{vs.\ }                 % against
\def\dt{\Delta t}
\def\obj{p_\text{obj}}
\def\iou{p_\text{iou}}

\def\cyc{two-wheeler\xspace}

\def\cycs{two-wheelers\xspace}
\def\Cycs{Two-Wheelers\xspace}

\definecolor{LavenderBlue}{rgb}{0.7020,    0.8039,    0.8902}
\definecolor{Lightapricot}{rgb}{0.9961,    0.8510,    0.6510}
\definecolor{thirdtablecolor}{rgb}{0.8706,    0.7961,    0.8941}

\newcommand{\heading}[1]{\noindent\textbf{#1}}
\newcommand*{\circled}[1]{\raisebox{.5pt}{\textcircled{\raisebox{-.9pt} {\footnotesize #1}}}}

\long\def\ignorethis#1{}

\definecolor{demphcolor}{RGB}{100,100,100}

\newlength\pagetopmargin
\newlength\figcapmargin
\newlength\figmargin
\newlength\tablecapmargin
\newlength\tablemargin

\usepackage{soul}
\newcommand{\algoNameFull}{LEOD\xspace}

\definecolor{cvprblue}{rgb}{0.21,0.49,0.74}
\usepackage[pagebackref,breaklinks,colorlinks,citecolor=cvprblue]{hyperref}

\usepackage[capitalize]{cleveref}
\crefname{section}{Sec.}{Secs.}
\Crefname{section}{Section}{Sections}
\Crefname{table}{Table}{Tables}
\crefname{table}{Tab.}{Tabs.}
\Crefname{figure}{Figure}{Figures}
\crefname{figure}{Fig.}{Figs.}

\def\paperID{3773} % *** Enter the Paper ID here
\def\confName{CVPR}
\def\confYear{2024}

\title{\algoNameFull: Label-Efficient Object Detection for Event Cameras}

\author{
Ziyi Wu$^{1,2}$, Mathias Gehrig$^3$, Qing Lyu$^1$, Xudong Liu$^1$, Igor Gilitschenski$^{1,2}$ \\
$^1$University of Toronto, $^2$Vector Institute, $^3$ University of Zurich \\
}

\begin{document}
\maketitle

\begin{abstract}
Object detection with event cameras benefits from the sensor's low latency and high dynamic range.
However, it is costly to fully label event streams for supervised training due to their high temporal resolution.
To reduce this cost, we present \algoNameFull, the first method for label-efficient event-based detection.
Our approach unifies weakly- and semi-supervised object detection with a self-training mechanism.
We first utilize a detector pre-trained on limited labels to produce pseudo ground truth on unlabeled events. Then, the detector is re-trained with both real and generated labels.
Leveraging the temporal consistency of events, we run bi-directional inference and apply tracking-based post-processing to enhance the quality of pseudo labels.
To stabilize training against label noise, we further design a soft anchor assignment strategy.
We introduce new experimental protocols to evaluate the task of label-efficient event-based detection on Gen1 and 1Mpx datasets.
\algoNameFull consistently outperforms supervised baselines across various labeling ratios.
For example, on Gen1, it improves mAP by 8.6\% and 7.8\% for RVT-S trained with 1\% and 2\% labels.
On 1Mpx, RVT-S with 10\% labels even surpasses its fully-supervised counterpart using 100\% labels.
\algoNameFull maintains its effectiveness even when all labeled data are available, reaching new state-of-the-art results.
Finally, we show that our method readily scales to improve larger detectors as well.
Code: \url{https://github.com/Wuziyi616/LEOD}.
\end{abstract}

\section{Introduction}

Object detection is key to scene understanding.
It provides a compact representation of raw sensor measurements as semantically meaningful bounding boxes.
Speed is crucial in object detection, especially in safety-critical applications such as self-driving.
Recently, event cameras have gained significant interest in computer vision due to their low latency, low energy consumption, and high dynamic range~\cite{EventVisionSurvey}.
Leveraging these benefits, event-based object detectors~\cite{AEGNN,1MpxDet,RVT,HMNet,ASTMNet,GWDDet} have been developed to complement conventional frame-based detectors.
Despite tremendous progress, much of their success heavily relies on large datasets that are manually annotated.
However, due to the high temporal resolution of event data, labeling objects at every timestamp is impractical.
For example, Gen1 dataset~\cite{Gen1Det} provides object labels at lower than 4 Hz.
As a result, existing methods only train their models on labeled events, and discard the remaining unlabeled data.
In contrast, we view this task as a weakly-supervised learning problem, which calls for detection methods that can leverage a mix of labeled and unlabeled events during training.

\begin{figure}[t]
    \vspace{\pagetopmargin}
    \vspace{1mm}
    \centering
    \includegraphics[width=0.99\linewidth]{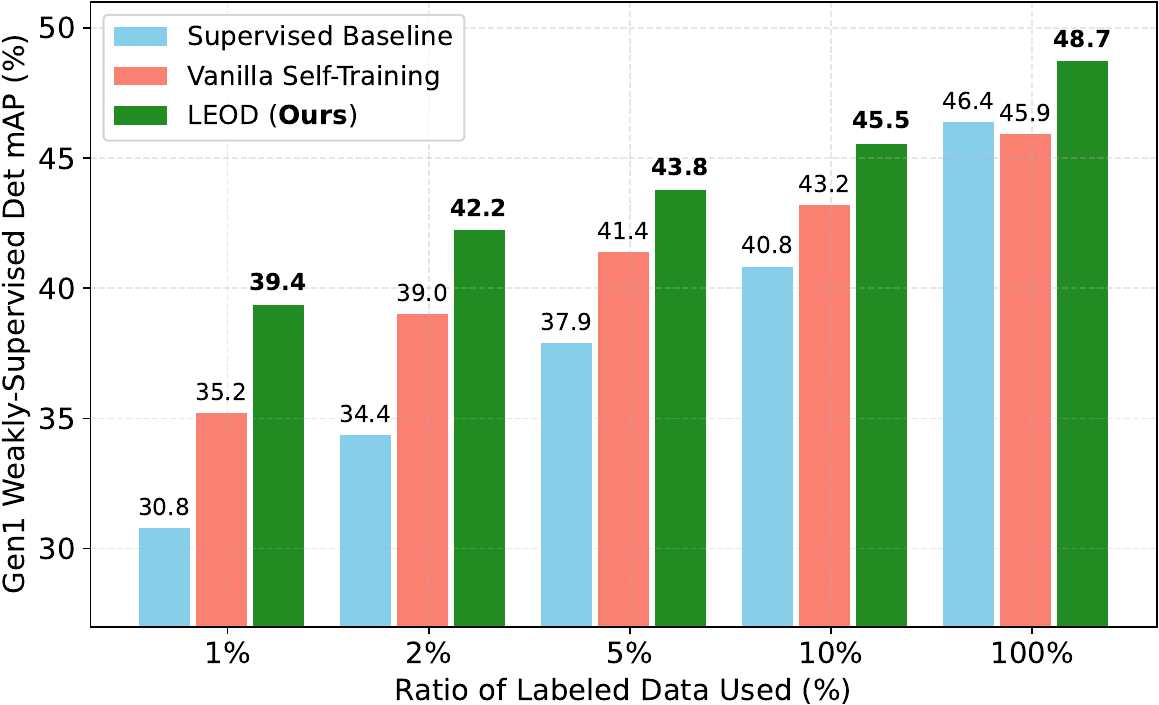}
    \vspace{\figcapmargin}
    \caption{
        \textbf{Detection performance} of \algoNameFull and baselines trained only on labeled events or conducting naive self-training.
        Under the weakly-supervised setting, our method consistently improves the RVT-S detector~\cite{RVT} across all labeling ratios on the Gen1 dataset.
    }
    \label{fig:teaser}
    \vspace{\figmargin}
\end{figure}

%
% Considered problem and general approach
%
In this work, we address the challenge by proposing a \textbf{L}abel-efficient \textbf{E}vent-based \textbf{O}bject \textbf{D}etection (\algoNameFull) framework.
We consider two settings characterized by limited labels:
\textbf{(i)} weakly-supervised, where object bounding boxes are sparsely labeled in all event streams, and
\textbf{(ii)} semi-supervised, where some event streams have object boxes densely labeled while others remain fully unlabeled.
Our approach unifies the two settings through a self-training paradigm.
With limited labels, we first pre-train a detector and use it to generate pseudo annotations on unlabeled events.
Then, we re-train the detector on a combination of real and pseudo labels.
However, naïvely generated labels contain noise, making their direct use suboptimal.

%
% Dealing with label noise
%
To obtain high-quality pseudo labels, we exploit the temporal dimension of event data.
Recent work~\cite{ASTMNet} has shown the importance of temporal information in event-based detection using recurrent modules~\cite{ConvLSTM}.
Additionally, offline label generation enables us to further refine predictions with future information.
To achieve this, we first introduce time-flip augmentation to events during training.
As a result, we can ensemble model predictions on the original and reverse event streams via Test-Time Augmentation (TTA), leading to higher detection recall.
In addition, we leverage tracking-based post-processing to eliminate temporally inconsistent objects, enhancing the precision of pseudo labels.
Finally, we filter out low-confidence boxes with a score threshold.

%
% Threshold selection
%
A key challenge here is how to select a proper threshold.
Instead of searching for the optimal value, we first filter with a low threshold to avoid missing objects.
This inevitably introduces many false positives, which we address with a soft anchor assignment strategy in training.
When computing the detection loss, we set another higher threshold and only treat pseudo boxes above that threshold as positive labels.
For boxes with a lower detection score, we ignore the loss applied to their associated anchors.
This strategy ensures that the model is only supervised with reliable background and foreground labels, while being tolerant to noisy labels.
Ablation studies show that our method is insensitive to the two threshold values, easing the hyper-parameter tuning.

%
% New Evaluation Protocols
%
To test our method, we design new protocols for label-efficient event-based detection on Gen1~\cite{Gen1Det} and 1Mpx~\cite{1MpxDet} datasets.
For weakly-supervised object detection (WSOD), we uniformly sub-sample the labels over time to simulate sparse annotations.
For semi-supervised object detection (SSOD), we directly choose some event sequences as fully unlabeled.
Following 2D SSOD evaluation protocols~\cite{STAC,SoftTeacher}, we also have a fully labeled setting, where we show that pseudo labels can complement ground-truths.

In summary, this work makes four main contributions:
\textbf{(i)} We introduce the task of label-efficient object detection to event vision, and design its experimental protocols.
\textbf{(ii)} We propose \algoNameFull, a unified framework for training event-based detectors with limited annotations.
\textbf{(iii)} \algoNameFull consistently outperforms baselines in various settings on two public detection datasets.
\textbf{(iv)} Our method remains effective under the fully labeled setting and scales up to larger detectors, achieving new state-of-the-art performance.

\section{Related Work}

\heading{Object Detection with Event Cameras.}
Existing event-based detectors can be mainly categorized into two classes depending on whether they utilize the asynchronous nature of events.
One line of work explores the sparsity of events, and employs Graph Neural Networks (GNNs)~\cite{AEGNN,EventSparseConv,EAGR} or Spiking Neural Networks (SNNs)~\cite{SNNEvDet1,SNNEvDet2,SNNEvDet3} for feature extraction.
However, these approaches struggle with propagating information over long time horizons, which is crucial for detecting objects from events.
Moreover, specialized hardware is required to achieve theoretical speed-ups of sparse networks, limiting their application in practice.

In another class of methods, events are converted to dense frame-like representations, followed by conventional networks for detection.
Earlier works only consider event frames aggregated from a short time interval~\cite{EVCNNDet1,EVCNNDet2,EVCNNDet3,AED}.
This discards long-horizon history and makes it hard to detect objects under small relative motion as they trigger very few events.
Recent methods thus introduce recurrent modules~\cite{LSTM,ConvLSTM} to enhance the memory of the detectors~\cite{1MpxDet,ASTMNet,RVT}.
Further research focuses on better backbones~\cite{ViT,MaxViT}, inference speed~\cite{HMNet}, and event representations~\cite{GWDDet}.
Because our primary goal is to study label-efficient learning for event-based detectors, we adopt the state-of-the-art approach RVT~\cite{RVT} as the base model.

\heading{Label-Efficient Learning in Event-based Vision.}
Due to a lack of large labeled datasets, there have been several works studying event-based algorithms with limited labels.
Some papers focus on bridging frame-based and event-based vision.
They either reconstruct natural images from events to apply traditional deep models~\cite{E2VID,E2VID-PAMI,FireNet,BetterE2VID,E2VID-Transformer}, simulate events from videos to transfer the annotations~\cite{DAVIS-Sim,ESIM,V2E,EventGAN}, or distill knowledge from trained frame-based models~\cite{Evdistill,ESS_EvSegFromImg,EventDA,WormholeLearningRSS,EventGraftNet,EvDataPretrain}.
However, these methods require either paired recordings of events and images or massive in-domain labeled images for training.
Closer to ours are methods that only use event data~\cite{MEM,EventCLIP,Ev-LaFOR,UnsupEvOptFlow}.
They conduct label-efficient learning on events with pre-trained frame-based models or self-supervised losses.
Yet, none of them are designed for the detection task.
Our work is the first attempt at label-efficient event-based object detection.

\heading{Label-Efficient Learning in Other Fields.}
Self-training based methods have been explored in tasks such as 2D image classification~\cite{SSCls1,FixMatch,NoisyStudent}, object detection~\cite{STAC,SoftTeacher,WeaklySup2DDet1}, and segmentation~\cite{WeaklySupSeg1,WeaklySupSeg2,WeaklySupSeg3}.
Our method is more related to label-efficient learning on videos~\cite{SemiSupVidDet1,SemiSupVidDet2,WeaklySupVidDet1} and 3D point cloud sequences~\cite{3DLiDARAutoLabeling,3DLiDARLESS,Oyster} as these methods also exploit the temporal information of input data.
For example, \cite{SemiSupVidDet2} utilizes optical flows to propagate single-frame labels to adjacent video frames.
\cite{3DLiDARAutoLabeling} and \cite{3DLiDARLESS} train a teacher model on dense point clouds aggregated from a few timesteps.
In contrast, we leverage a much longer temporal horizon by running the detector on the entire event stream in both directions.
\cite{Oyster} also employs tracking-based post-processing to remove inconsistent boxes.
We additionally perform tracking in both directions as a forward-backward consistency check, thus better leveraging the temporal information of event data.
To tackle the noisy pseudo labels, \cite{SoftTeacher} proposes to use the detection scores from the teacher model to weigh the loss.
Instead, we design a soft anchor assignment strategy by ignoring the loss associated with unconfident boxes.

\section{Method}

This paper introduces a new task named label-efficient event-based object detection (formulated in \cref{sec:problem-formulation}).
Our algorithm adopts a two-stage self-training framework with reliable label selection (\cref{sec:leod-overview}), where we leverage the temporal information of event data to obtain high-quality pseudo labels and suppress noisy predictions (\cref{sec:pseudo-label-filtering}).

\subsection{Problem Formulation}\label{sec:problem-formulation}

\heading{Events data.}
Event cameras record brightness changes, and output a sequence of events $\mathcal{E} = \{e_i = (x_i, y_i, t_i, p_i)\}$.
Each event $e_i$ is parameterized by its pixel coordinate $(x_i, y_i)$, timestamp $t_i$, and polarity $p_i \in \{-1, 1\}$.
Modern event cameras run at sub-milliseconds and can produce millions of events per second~\cite{EventVisionSurvey}.

\heading{Event-based Object Detection.}
As the object motion in a scene is usually much slower than the event generation speed, event-based object detectors are only applied and evaluated at a fixed time interval $T$~\cite{Gen1Det,1MpxDet}.
More specifically, given an event stream $\mathcal{E}$ capturing a set of objects $\mathcal{O} = \{o_j\}_{j=1}^M$, we aim at detecting the 2D bounding boxes of them with the semantic labels, $\mathcal{B} = \{b_j = (x_j, y_j, w_j, h_j, l_j, t_l)\}_{j=1}^M$.
Each bounding box $b_j$ is characterized by the location of its top-left corner $(x_j, y_j)$, width $w_j$, height $h_j$, class label $l_j \in \{1, 2, ..., C\}$, and timestamp $t_j$.
Here, $C$ denotes the number of classes.

\heading{Event-based Object Detectors.}
We take RVT~\cite{RVT} for example as it serves as our base model.
RVT is a synchronous detector that converts events in every time window $\dt$ to a grid-like representation $I$.
In the remaining part of the paper, we call $I$ a \emph{frame} and every $\dt$ a \emph{timestep}.
RVT combines a Vision Transformer backbone~\cite{MaxViT} with a YOLOX detection head~\cite{YOLOX}.
To extract temporal features, RVT introduces LSTM~\cite{LSTM} cells in the backbone to fuse information over multiple timesteps.
This is useful for detecting slow-moving objects as they only generate a few events.
The YOLOX detection head in RVT is anchor-free, \ie for each location on the feature map (anchor point), it predicts an objectness score $\obj \in [0, 1]$, per-class IoU values $\iou \in \mathbb{R}^C$, and offsets of the bounding box parameters $\Delta b = (\Delta x, \Delta y, \Delta w, \Delta h)$.
The $\iou$ is trained to output the IoU value between the predicted box and the matched ground-truth box of that class.
To obtain the final prediction, Non-Maximum Suppression (NMS) is applied to remove overlapping bounding boxes with low confidence.

During training, each ground-truth bounding box is matched to several anchor points for loss computation.
Below we will denote anchor points as \emph{anchors} for simplicity.
Anchors matched with ground-truth are foreground, where all predicted values are supervised.
For the remaining background anchors, only $\obj$ is trained to be $0$.

\begin{figure}[t]
    \vspace{\pagetopmargin}
    \vspace{-3.5mm}
    \centering
    \begin{subfigure}{0.99\linewidth}
        \includegraphics[width=1.0\linewidth]{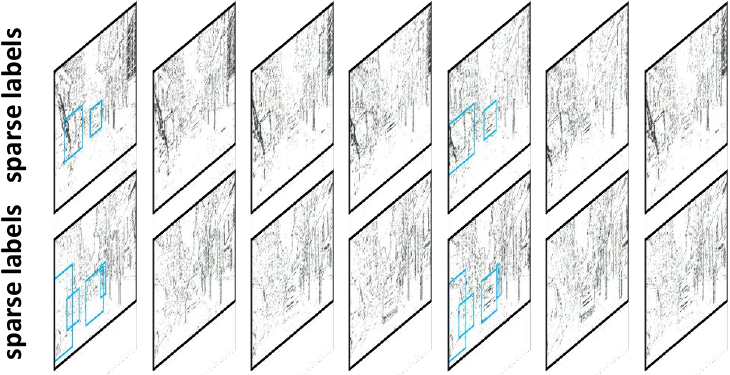}
        \vspace{-5mm}
        \caption{Weakly-supervised object detection (WSOD)}
    \end{subfigure}
    \vspace{1.2mm}\\
    \begin{subfigure}{0.99\linewidth}
        \includegraphics[width=1.0\linewidth]{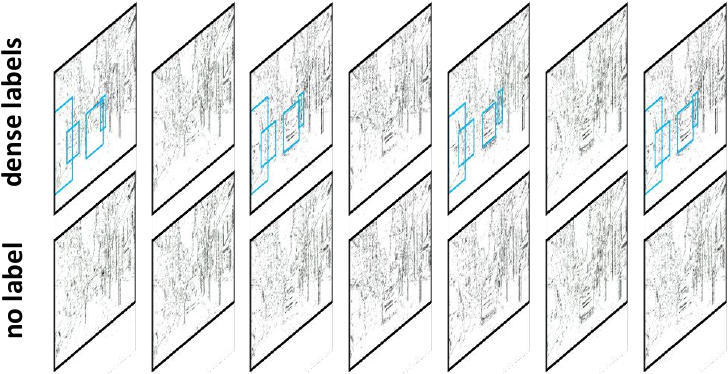}
        \vspace{-5mm}
        \caption{Semi-supervised object detection (SSOD)}
    \end{subfigure}
    \vspace{-2.5mm}
    \caption{
        \textbf{Illustration of two label-efficient event-based object detection settings}: (a) weakly-supervised where all event sequences are sparsely annotated, and (b) semi-supervised where some event sequences are densely annotated, and others are fully unlabeled.
        We visualize both positive and negative events in black. %in this paper.
    }
    \label{fig:ssod-wsod-settings}
    \vspace{\figmargin}
\end{figure}

\heading{Label-Efficient Event-based Object Detection.}
\cref{fig:ssod-wsod-settings} shows our proposed detection settings with limited labels.

\noindent
\textit{Weakly-supervised object detection (WSOD).}
In the WSOD setting, all event streams are sparsely labeled.
Moreover, labels assigned to adjacent frames offer fewer informative training signals compared to those distributed across frames~\cite{SemiSupVidDet2}.
Therefore, it is reasonable to label object boxes in a long event sequence uniformly and sparsely.

\noindent
\textit{Semi-supervised object detection (SSOD).}
In the SSOD setting, some event sequences are densely labeled, while others are fully unlabeled.
This is also practical when people continue to collect data into already annotated datasets.
Since capturing event sequences is much easier than labeling them, an algorithm that can consistently improve model performance with raw events is highly useful.

To evaluate the label-efficient learning performance, we take existing event-based object detection datasets~\cite{Gen1Det,1MpxDet} and sample a small portion of frames (WSOD) or sequences (SSOD) as labeled data.
The rest of the training data are used as an unlabeled set following previous works~\cite{STAC,UnbiasedTeacher,SoftTeacher}.
We also have a fully labeled setting where all labels are available.
Since the original event streams are annotated at a larger time interval than $\dt$, we can still create pseudo labels on unlabeled timesteps to improve the performance.
See Appendix~\ref{app:discuss-wsod-ssod} for further discussions on the two settings.

\begin{figure*}[t]
    \vspace{\pagetopmargin}
    \vspace{-3.3mm}
    \centering
    \includegraphics[width=0.99\linewidth]{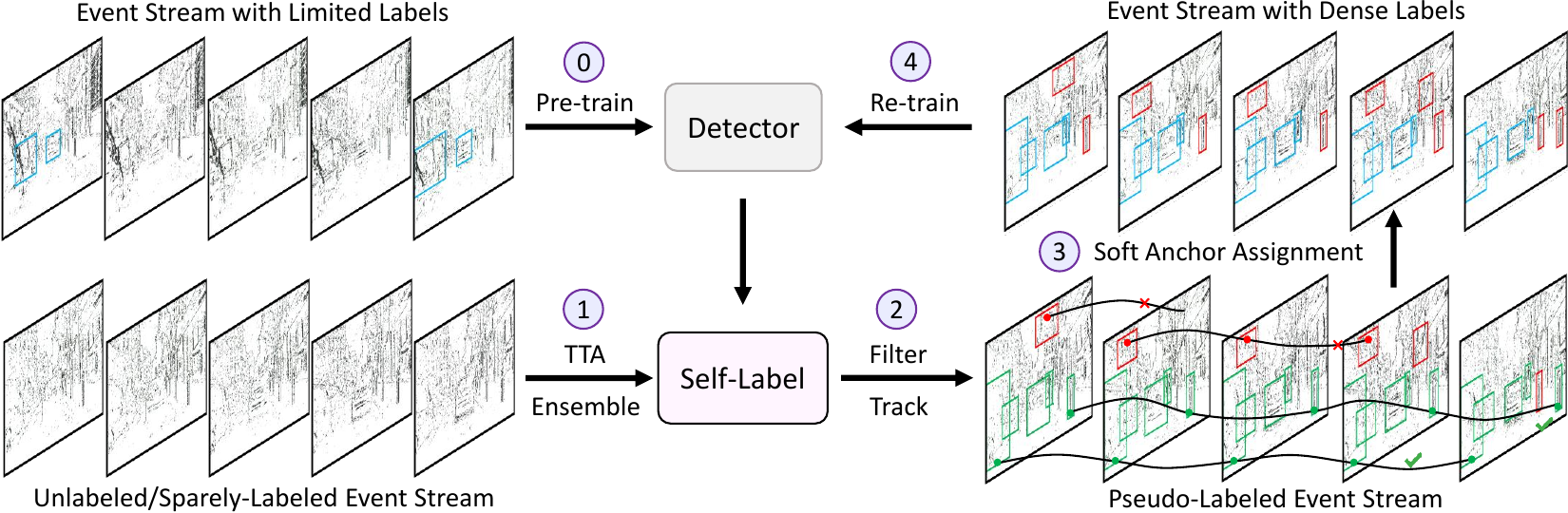}
    \vspace{\figcapmargin}
    \caption{
        \textbf{Overview of our \algoNameFull pipeline.}
        \circled{0} We first pre-train an event-based object detector on event streams with limited labels.
        \circled{1} To leverage the temporal information, we apply time-flip Test-Time Augmentation (TTA) to unlabeled event streams and ensemble the model predictions.
        \circled{2} We then apply forward and backward tracking to identify temporally inconsistent bounding boxes, \ie boxes associated with short tracks.
        \circled{3} To handle noisy labels, a soft anchor assignment strategy is designed to ignore detection loss on unconfident pseudo labels ({\color{red} red} boxes).
        \circled{4} We can boost the model performance by self-training on reliable pseudo labels ({\color{cyan} blue} boxes) and repeating \circled{1} -- \circled{4}.
    }
    \label{fig:pipeline}
    \vspace{\figmargin}
    \vspace{0.5mm}
\end{figure*}

\subsection{\algoNameFull: A Self-Training Framework}\label{sec:leod-overview}

As shown in \cref{fig:pipeline}, the overall pipeline of \algoNameFull follows a student-teacher pseudo-labeling paradigm, which is applicable to both, the WSOD and the SSOD settings.
We first pre-train a detector on labeled data using regular detection loss until convergence.
Then, we employ it to annotate unlabeled frames.
To leverage the benefit of offline prediction, we apply temporal flip to get event streams in both directions, and aggregate the detection results on them.
Since the teacher model is trained on limited data, it will be uncertain on hard examples.
We thus threshold the boxes with a small value to keep more detected objects.
To remove false positive boxes, we build upon the temporal persistency prior of objects and apply tracking-based post-processing.
However, there might still be inaccurate labels due to the low confidence threshold we use, and directly training the detector on them will lead to suboptimal results.
Inspired by prior works on noise-robust learning~\cite{DetwNoisyLabels,CutLer}, we design a soft anchor assignment strategy to selectively supervise the model with pseudo labels.
Finally, we can use the re-trained detector as the teacher model to initialize the next round of self-training.
The described process can be repeated for multiple rounds to further boost the model performance.

\heading{Comparison to online pseudo-labeling.}
In previous label-efficient object detection works~\cite{UnbiasedTeacher,SoftTeacher,3DDetProficientTeachers,3DIoUMatchDet}, the teacher model is jointly trained with a student model.
In each training step, the teacher predicts bounding boxes on unlabeled data in a batch for student training.
This online paradigm is also applicable to our setting.
However, the teacher model will only see short event streams loaded in a batch.
For example, on Gen1~\cite{Gen1Det}, our training sequence length accounts for a duration of 1 second, while a car can stop for more than 10 seconds in real-world traffic and thus trigger no events.
Pseudo-labeling on short event streams will inevitably miss these objects.
Therefore, we adopt our two-stage offline label generation paradigm to retain full temporal information.

\begin{figure}[t]
    % \vspace{-1mm}
    \centering
    \includegraphics[width=0.99\linewidth]{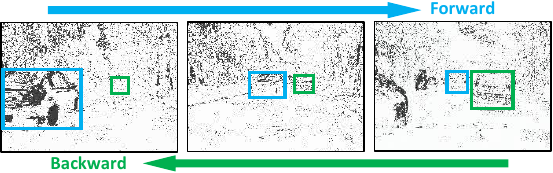}
    \vspace{\figcapmargin}
    \caption{
        \textbf{Illustration of the time-flip TTA} which enhances our robustness against different object motions.
        {\color{cyan} Forward} helps detect receding objects, while {\color{Green} Backward} helps with approaching objects.
    }
    \label{fig:tta-vis}
    \vspace{\figmargin}
\end{figure}

\subsection{Towards High-Quality Pseudo Labeling}\label{sec:pseudo-label-filtering}

In this section, we introduce each key component in our \algoNameFull framework to achieve high-quality pseudo labels.

\heading{Test-Time Augmentation (TTA).}
When deployed in the real world, event-based detectors are expected to run in real-time, \ie they only take in events triggered before $t$ to detect objects at $t$.
Instead, in our offline label generation process, we can use future information to refine predictions at the current timestep.
As shown in \cref{fig:tta-vis}, we run the detector on both the original and reversed event streams, enabling us to detect objects with different movements.
We also apply a horizontal-flip TTA to further improve the detection.

\begin{figure*}[t]
    \vspace{\pagetopmargin}
    \vspace{-3mm}
    \centering
    \begin{subfigure}{0.23\linewidth}
        \includegraphics[width=1.0\linewidth]{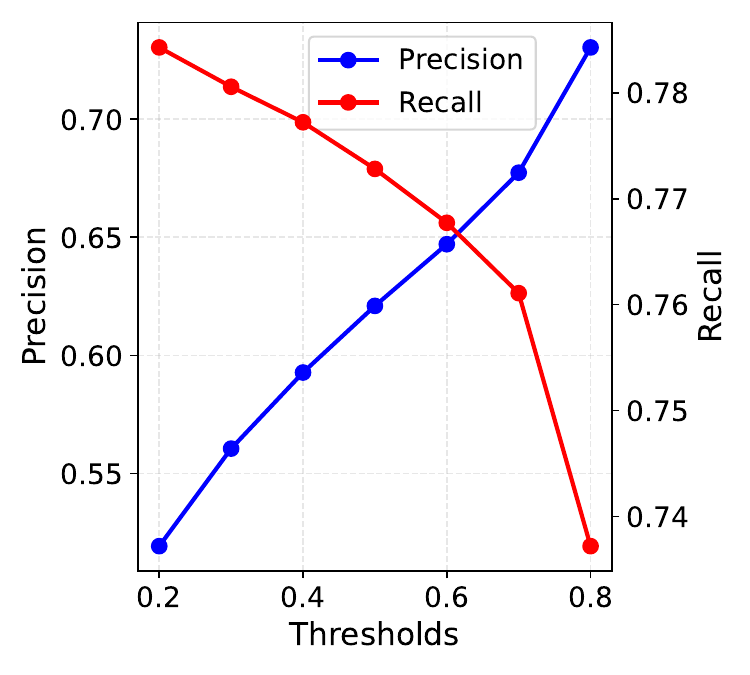}
        \vspace{-5mm}
        \caption{Label quality of cars}
    \end{subfigure}
    \hspace{-2.5mm}
    \begin{subfigure}{0.23\linewidth}
        \includegraphics[width=1.0\linewidth]{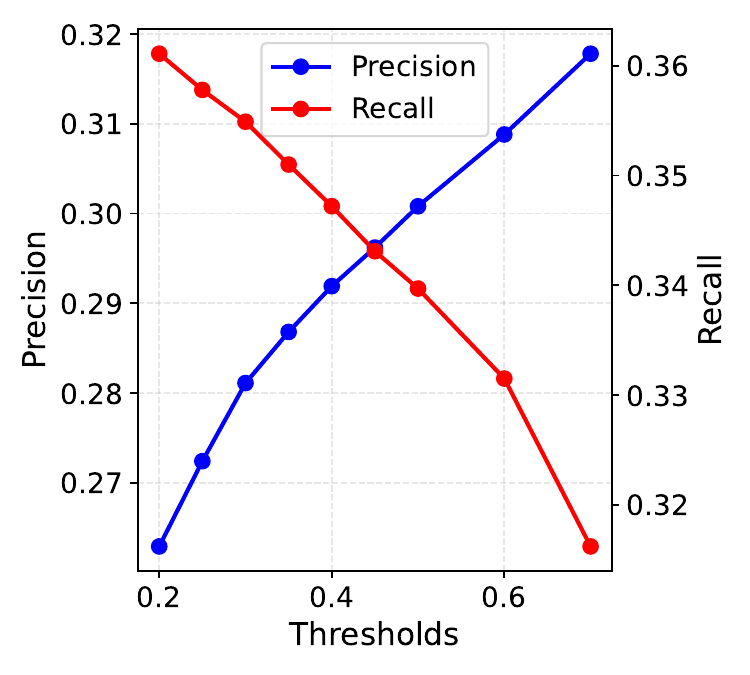}
        \vspace{-5mm}
        \caption{Label quality of pedestrians}
    \end{subfigure}
    \hspace{-2mm}
    \begin{subfigure}{0.272\linewidth}
        \includegraphics[width=1.0\linewidth]{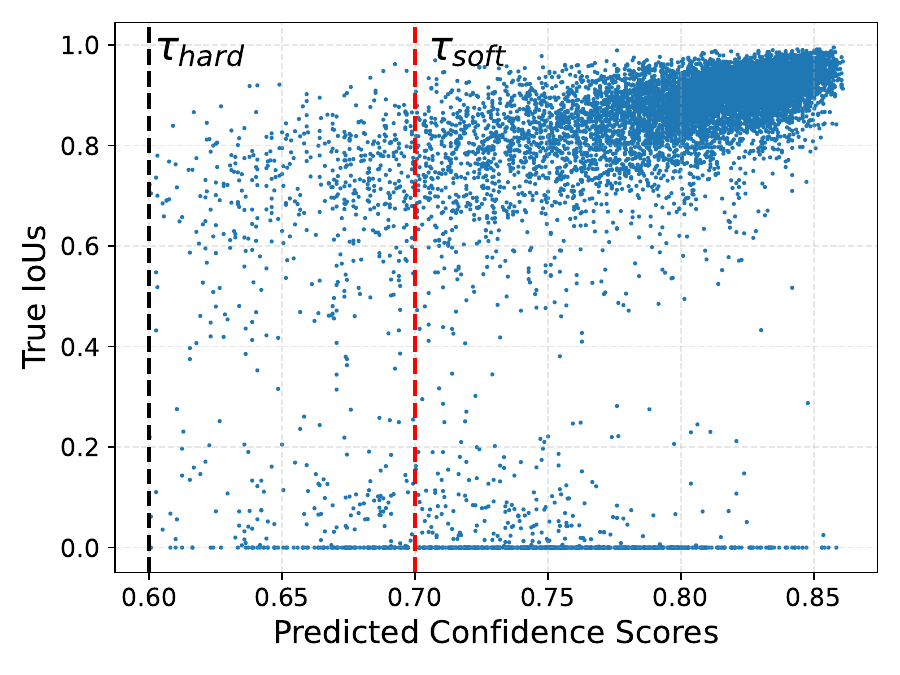}
        \vspace{-5mm}
        \caption{Scores \vs True IoUs of cars}
    \end{subfigure}
    \hspace{-2mm}
    \begin{subfigure}{0.272\linewidth}
        \includegraphics[width=1.0\linewidth]{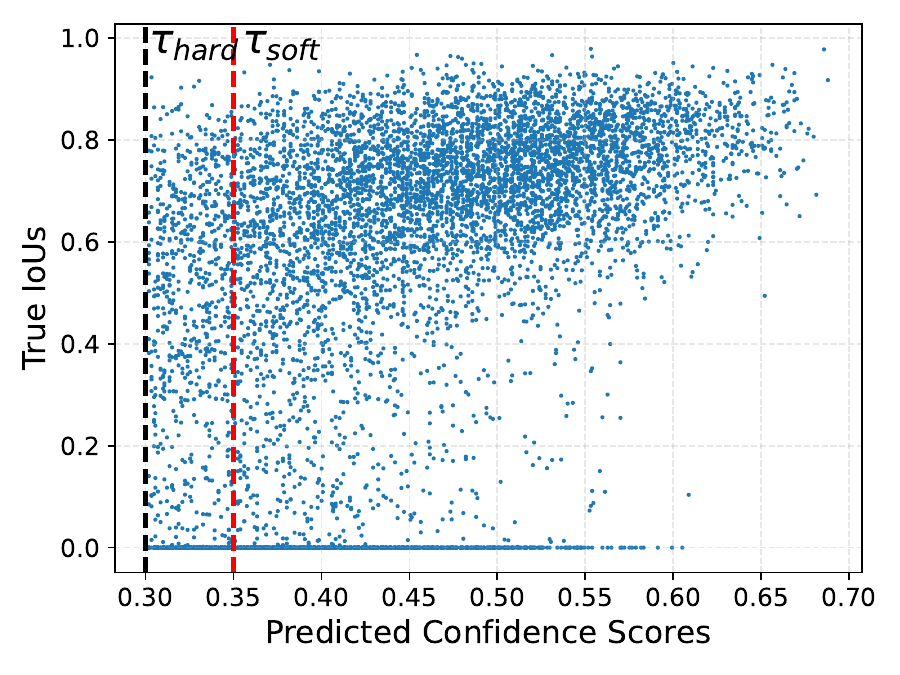}
        \vspace{-5mm}
        \caption{Scores \vs True IoUs of pedestrians}
    \end{subfigure}
    \vspace{-2.5mm}
    \caption{
        \textbf{Analysis on confidence thresholds.}
        We randomly sample 10,000 predicted boxes from RVT-S pre-trained on 5\% of Gen1 labels.
        We plot the pseudo labels' precision and recall of (a) cars and (b) pedestrians.
        In (c) and (d), we show each box's predicted confidence scores and its true IoU with ground-truth boxes.
        $\tau_\text{hard}$ is the threshold for initial filtering, and $\tau_\text{soft}$ is used in soft anchor assignment.
    }
    \label{fig:pr-curves}
    \vspace{\figmargin}
\end{figure*}

\heading{Filtering and Tracking.}
TTA helps us detect more objects (higher recall), yet it also leads to false positives (lower precision).
Previous works simply use a threshold to filter out boxes with low confidence~\cite{SoftTeacher,3DIoUMatchDet}.
However, as shown in \cref{fig:pr-curves} (a) and (b), there is a trade-off between precision and recall, making it hard to find the optimal threshold.
We opt to first filter with a low threshold $\tau_\text{hard}$ to avoid missing objects, followed by tracking-based post-processing to remove temporally inconsistent boxes.
We follow the tracking-by-detection paradigm~\cite{SORT} to build tracks by linking detection boxes between frames.
Each box $b$ will be associated with a track $s_k = \{(b, v_x, v_y)_t, k, n, q\}$, where $(v_x, v_y)$ is the estimated velocity under a linear motion assumption, $k$ is the ID, $n$ is its length so far, and $q \in [0, 1]$ is the current score.
In the first frame, we initialize each box as a track.
For every coming frame, we first predict the positions of existing tracks and associate them with boxes at that frame via greedy matching over pairwise IoUs.
Then, we decay the score $q$ of unmatched tracks and initialize unmatched boxes as new tracks.
Finally, tracks with low scores will be deleted.
See Appendix~\ref{app:tracking} for implementation details.

Similar to TTA, we apply tracking in both directions, and only remove a box if the length of its associated track is shorter than a threshold, $T_{trk}$, in both cases.
However, predictions on hard examples may also be inconsistent, as the pre-trained model has limited capacity.
Instead of suppressing removed boxes as background, we ignore them during loss computation as will be described later.
Also, for long tracks, we inpaint boxes with linear motion at unmatched timesteps and ignore training losses on them too.

\heading{Soft Anchor Assignment and Re-training.}
We can now re-train a detector on the ground-truth labels and the pseudo labels with the original detection loss.
However, as shown in \cref{fig:pr-curves} (c) and (d), there are still low-quality boxes after post-processing.
To handle noisy labels, we utilize a soft anchor assignment strategy to selectively supervise the model training.
We first identify a set of uncertain labels including boxes belonging to short tracks, inpainted from long tracks, and those with detection scores lower than a threshold $\tau_\text{soft}$.
Then, at each training step, we ignore the loss applied to anchors associated with these uncertain boxes, \ie we do not supervise those anchors and allow them to discover new instances.
This design is inspired by the anchor assignment in anchor-based detectors~\cite{Faster_RCNN,Retina_Net}, where two thresholds are used to determine foreground or background anchor boxes, and the anchors in between are ignored in loss computation.
As we will show in ablation studies, soft anchor assignment makes our method less sensitive to hyper-parameters.

Despite training on noisy labels, the model learns to refine the labels and detect new objects.
Thus, we do an additional round of self-training to further improve the results.

\section{Experiments}

\cref{sec:less-label-results} shows that \algoNameFull outperforms baselines significantly in the low-label regime.
In the fully labeled setting, we surpass previous state-of-the-art (\cref{sec:full-label-results}).
Our ablations show the contribution of each component in \cref{sec:ablation}.

\begin{figure*}[t]
    \vspace{\pagetopmargin}
    \vspace{-4mm}
    \centering
    \begin{subfigure}{0.49\linewidth}
        \includegraphics[width=1.0\linewidth]{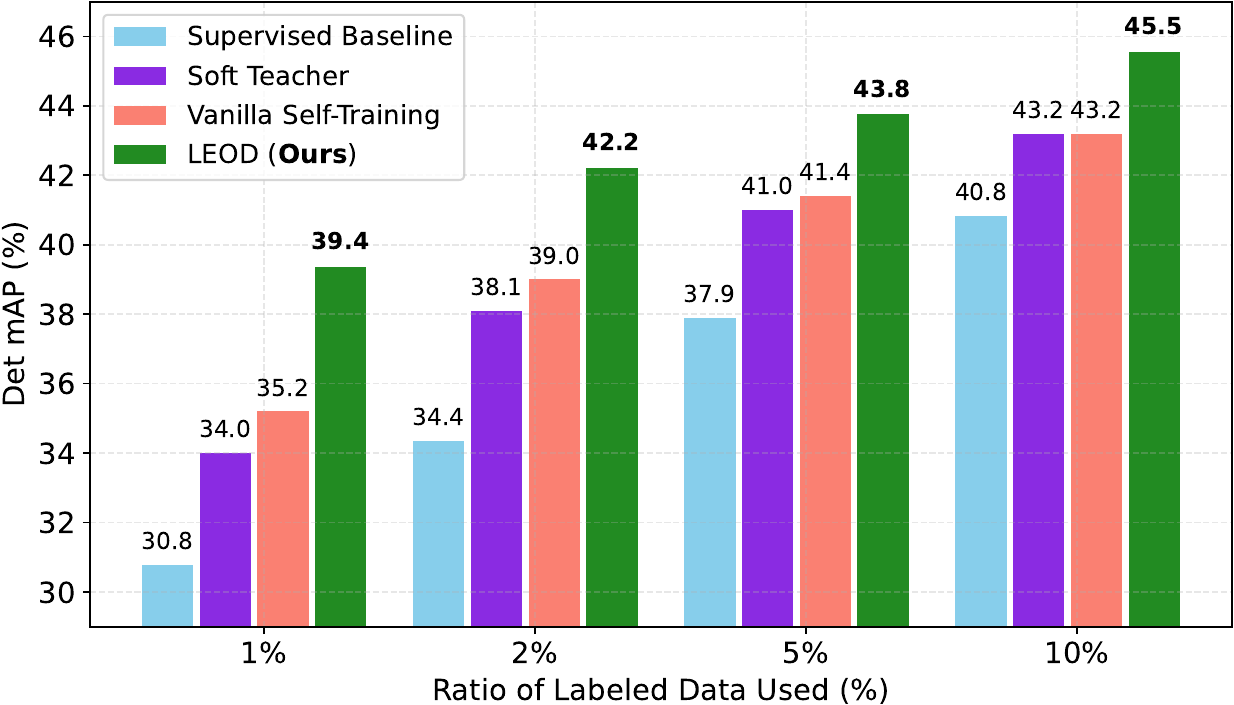}
        \vspace{-4mm}
        \caption{Gen1 weakly-supervised object detection (WSOD) results}
    \end{subfigure}
    \hspace{1mm}
    \begin{subfigure}{0.49\linewidth}
        \includegraphics[width=1.0\linewidth]{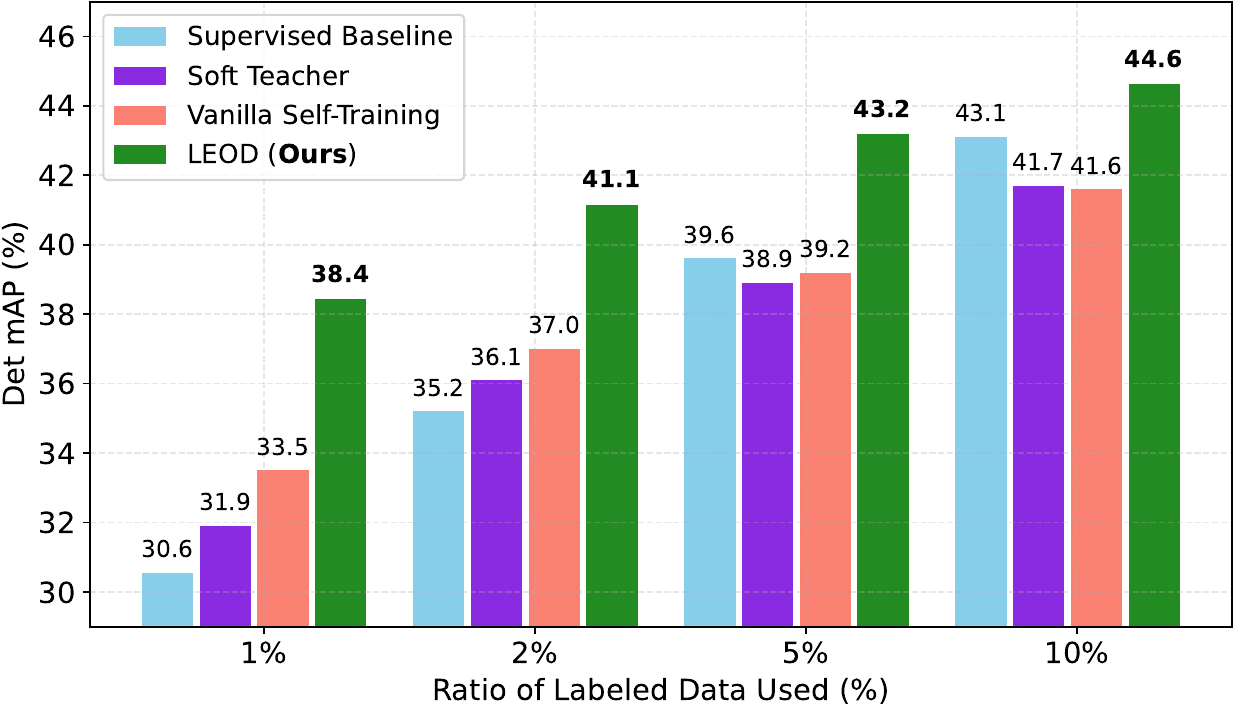}
        \vspace{-4mm}
        \caption{1Mpx weakly-supervised object detection (WSOD) results}
    \end{subfigure}
    \\
    \vspace{2mm}
    \begin{subfigure}{0.49\linewidth}
        \includegraphics[width=1.0\linewidth]{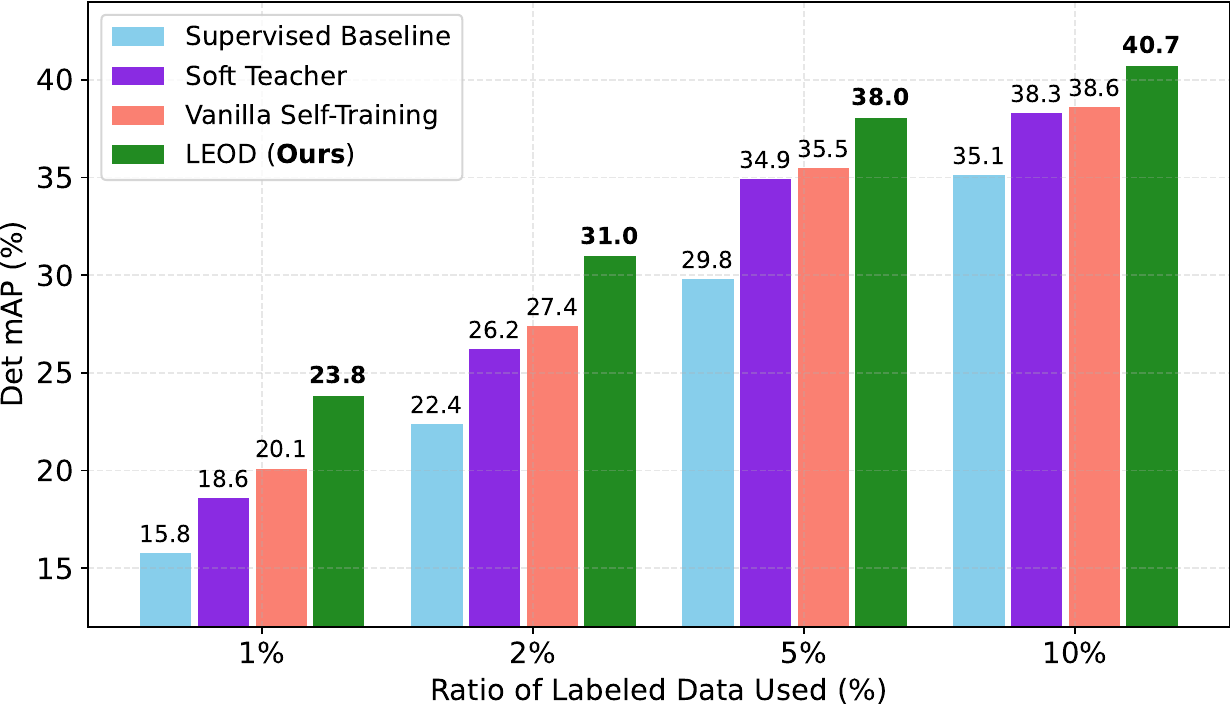}
        \vspace{-4mm}
        \caption{Gen1 semi-supervised object detection (SSOD) results}
    \end{subfigure}
    \hspace{1mm}
    \begin{subfigure}{0.49\linewidth}
        \includegraphics[width=1.0\linewidth]{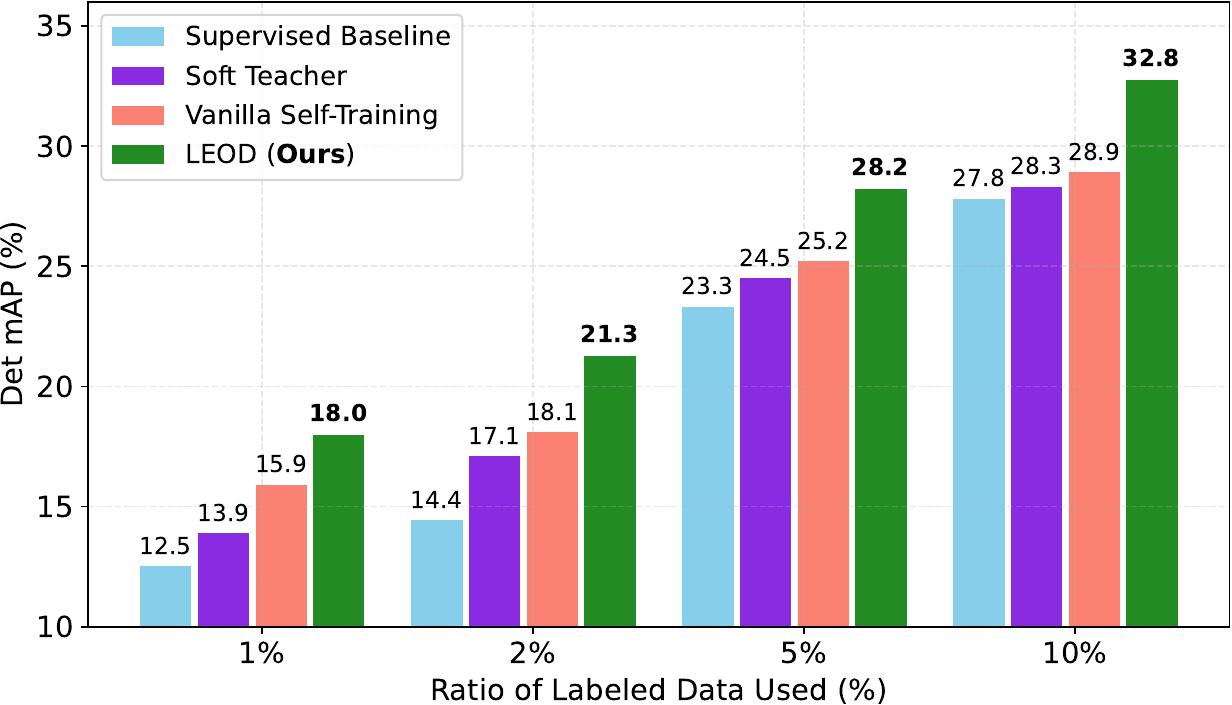}
        \vspace{-4mm}
        \caption{1Mpx semi-supervised object detection (SSOD) results}
    \end{subfigure}
    \vspace{-2mm}
    \caption{
        \textbf{Results on label-efficient learning} using different ratios of labeled data.
        We compare \algoNameFull with baselines on Gen1 and 1Mpx datasets under WSOD and SSOD using the RVT-S detector.
        All results are averaged over three runs.
    }
    \label{fig:limited-label-result}
    \vspace{\figmargin}
\end{figure*}

\subsection{Experimental Setup}\label{sec:exp-setup}

\heading{Datasets.}
We adopt Gen1~\cite{Gen1Det} and 1Mpx~\cite{1MpxDet} datasets that feature various driving scenarios.
Gen1 consists of 39 hours of recordings with a 304$\times$240 resolution event camera~\cite{Gen1_camera}.
It provides bounding box annotations of cars and pedestrians at 1, 2, or 4 Hz.
1Mpx is recorded with a higher 720$\times$1280 resolution event camera~\cite{Gen4_camera}.
It contains around 15 hours of data collected over several months at day and night, and provides labels for cars, pedestrians, and \cycs at 30 or 60 Hz.
Following previous works~\cite{ASTMNet,RVT}, we remove ground-truth boxes that are too small during evaluation on both datasets, and half the events' resolution to 360$\times$640 on 1Mpx.

\heading{Evaluation Protocol.}
Mean average precision (mAP) is adopted as the main performance metric.
We choose 1\%, 2\%, 5\%, and 10\% as the labeling ratio following prior works~\cite{STAC,SoftTeacher}.
In the weakly-supervised object detection (WSOD) setting, labels in all event streams are uniformly sub-sampled.
In the semi-supervised object detection (SSOD) setting, we keep a small portion of event streams unchanged, while setting other event sequences as fully unlabeled.
For the same labeling ratio, the amounts of available labels in WSOD and SSOD are roughly the same.
Finally, all labels are provided in the fully labeled setting.

\heading{Detector Training Details.}
We adopt the state-of-the-art event-based detector RVT~\cite{RVT} as our base model.
Due to limited computation resources, we mainly experiment with RVT-S, while we show that \algoNameFull also scales to the largest RVT-B variant in \cref{sec:ablation}.
Most of the configurations are the same as we build upon their open-source codebase.
Here we only highlight our modifications.
In order to apply the time-flip TTA, we train with an additional time-flip data augmentation.
When re-training on pseudo labels, we initialize RVT from scratch and use the Adam optimizer~\cite{Adam} with a peak learning rate of $5\times10^{-4}$ to train for 150k iterations.
Please refer to Appendix~\ref{app:rvt-training} for more details.

\heading{Pseudo-Labeling Details.}
Inspired by prior works~\cite{PV-RCNN,3DIoUMatchDet}, we set different thresholds for each category.
To simplify parameter tuning, we follow two rules:
\textbf{(i)} pedestrians and \cycs share the same values, which are half of cars' values,
\textbf{(ii)} for cars, the soft threshold $\tau_\text{soft}$ equals to the hard threshold $\tau_\text{hard} + 0.1$, while for pedestrians and \cycs, we use $\tau_\text{soft} = \tau_\text{hard} + 0.05$.
In both settings and both datasets, we choose the same set of hyper-parameters: $\tau_\text{hard} = 0.6$ for cars and $\tau_\text{hard} = 0.3$ for pedestrians.
Only in 1Mpx WSOD, we set $\tau_\text{hard} = 0.5$ for pedestrians and \cycs to handle excessive false positives.
The minimum track length $T_{trk}$ is set to 6 in all experiments.

\heading{Baselines.}
We compare with a \textit{Supervised Baseline} trained only on available labels.
Since we are the first work to consider this task, we design two other baselines: \textit{1) Vanilla Self-Training:} trains on pseudo labels without TTA, tracking, and soft anchor assignment; \textit{2) Soft Teacher:} adopts the online student-teacher paradigm from a representative 2D SSOD method~\cite{SoftTeacher}.
We enable soft anchor assignment in \textit{Soft Teacher}, while TTA and tracking are not applicable as the online event sequence length is too short.
We tune baselines' hyper-parameters to be optimal in each setting.
We also tried a state-of-the-art 2D SSOD method designed for anchor-free detectors~\cite{UnbiasedTeacherV2} as YOLOX is anchor-free, but we did not observe clear improvements over Soft Teacher.

% qualitative detection results
\begin{figure*}[t]
    \vspace{\pagetopmargin}
    \vspace{-2mm}
    \centering
    \includegraphics[width=0.98\linewidth]{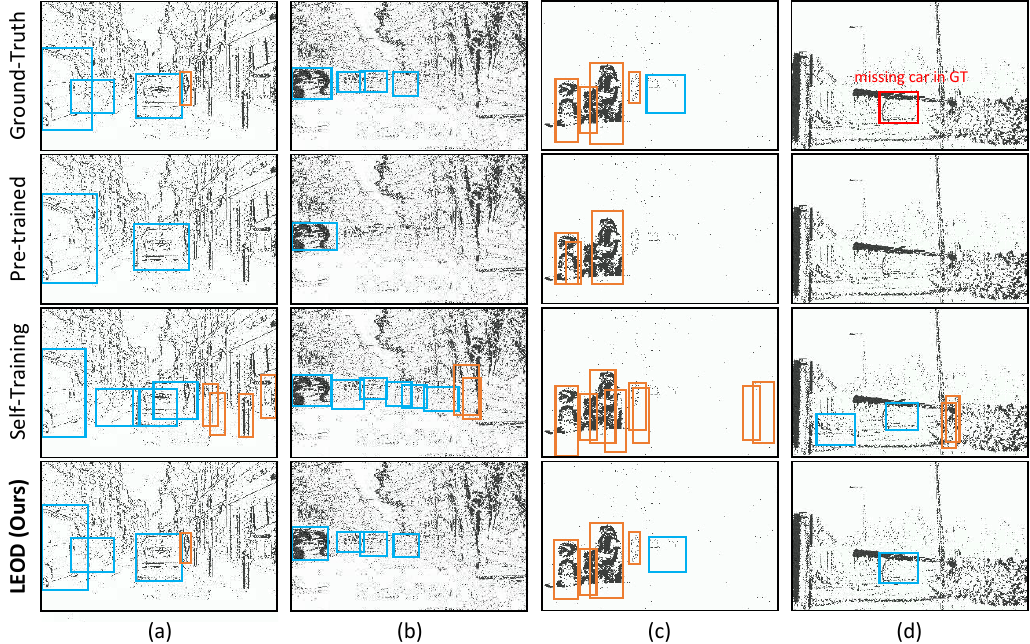}
    % \vspace{-1mm}
    \vspace{\figcapmargin}
    \caption{
        \textbf{Detection on Gen1.}
        We show (a) a common driving scene, (b) cars with small relative movement thus few events, (c) crowded pedestrians, and (d) \algoNameFull discovers a missing object in ground-truth annotations due to the occlusion of a barrier gate.
        Car and pedestrian boxes are colored in {\color{cyan} blue} and {\color{orange} orange}, respectively.
        Pre-trained and Self-Training stand for \textit{Supervised Baseline} and \textit{Vanilla Self-Training}.
    }
    \label{fig:qual-det-result}
    \vspace{\figmargin}
\end{figure*}

\subsection{Label-Efficient Results}\label{sec:less-label-results}

We first compare our method with baselines in the low-labeled data regime.
The overall results are shown in \cref{fig:limited-label-result}.

\heading{WSOD.}
\cref{fig:limited-label-result} (a) and (b) present the weakly-supervised results.
On Gen1, \algoNameFull improves the mAP of Supervised Baseline by a large margin across all labeling ratios.
Using 10\% labels, we achieve an mAP of 45.5\%, which is only 1\% lower than RVT-S trained on 100\% labels.
For pseudo-labeling baselines, Vanilla Self-Training outperforms Soft Teacher in most cases, validating our choice of offline label generation.
In addition, \algoNameFull consistently outperforms them by more than 2\%, indicating the higher quality of our pseudo labels.
We observe a similar trend on 1Mpx, where our approach scores the highest mAP in all cases.
Notably, the two pseudo-labeling baselines perform worse than pre-trained RVT-S on 5\% and 10\% labels, which proves the importance of screening reliable labels.
Finally, \algoNameFull with 10\% labels (44.6\%) outperforms RVT-S trained on all labels (44.1\%), showing the great potential of unlabeled data.

\heading{SSOD.}
\cref{fig:limited-label-result} (c) and (d) present the semi-supervised results.
Using the same amount of labels, models trained under SSOD are generally worse than WSOD.
This indicates that given a limited budget, we should sparsely label as many event streams as possible instead of densely labeling a few sequences.
Nevertheless, \algoNameFull still outperforms baselines by more than 2\% mAP over all labeling ratios on both datasets.
Our results offer a promising direction of boosting performance with fully unlabeled event data.

\heading{Qualitative Results.}
\cref{fig:qual-det-result} visualizes some detection results of RVT-S trained with 10\% labels under Gen1 WSOD setting.
The supervised baseline can only detect objects that trigger lots of events due to its limited capacity.
Models using vanilla self-training detect more objects, but also produce numerous false positives.
With our pseudo-labeling pipeline, \algoNameFull trained models are able to handle various hard examples.
For example, it discovers a car that was initially missed in the ground-truth annotation in \cref{fig:qual-det-result} (d).

\subsection{Fully-Labeled Results}\label{sec:full-label-results}

Since the original labeling frequency on both datasets is lower than the frame rate of RVT-S (20 Hz), we can still create pseudo labels on unlabeled frames to improve the fully supervised model performance.
We compare with state-of-the-art event-based object detectors trained on all labeled data in \cref{tab:full-label-result}.
\algoNameFull improves over RVT-S by 2.2\% and 2.6\% on Gen1 and 1Mpx, respectively.
On Gen1, our method achieves new state-of-the-art among models not using pre-trained weights.
This indicates that \algoNameFull is consistently effective even with 100\% labels.
In terms of runtime and model size, since our approach does not introduce new modules to the base model, we are as efficient as RVT-S.

% 100\% labels result
\begin{table}[t]
    \vspace{2mm}
    \centering
    \setlength{\tabcolsep}{4pt}
    \scriptsize
        \rowcolors{3}{uoftgray!50}{white}
    \begin{tabular}{lcc|ccc}
        \toprule
        \multirow{2}{*}{\textbf{Method}} & \multicolumn{2}{c}{Gen1} & \multicolumn{2}{c}{1Mpx} & \multirow{2}{*}{Size (M)} \\
        % \midrule
         & mAP (\%) & Time (ms) & mAP (\%) & Time (ms) & \\
        \midrule
        RED~\cite{1MpxDet} & 40.0 & 16.7 & 43.0 & 39.3 & 24.1 \\
        ASTMNet~\cite{ASTMNet} & 46.7 & 35.6 & \textbf{48.3} & 72.3 & $>$ 100 \\
        HMNet-L3~\cite{HMNet} & 47.1 & 7.9$^*$ & - & - & 33.2 \\
        RVT-B & 47.2 & 10.2 & 47.4 & 11.9 & 18.5 \\
        \midrule
        RVT-S & 46.5 & 9.5 & 44.1 & 10.1 & 9.9 \\
        \textbf{\algoNameFull-RVT-S} & \textbf{48.7} & 9.5 & 46.7 & 10.1 & 9.9 \\
        \midrule
        \color{gray} ERGO-12~\cite{GWDDet} & \color{gray} 50.4 & \color{gray} 77.2 & \color{gray} 40.6 & \color{gray} 101.1 & \color{gray} 59.6 \\
        \bottomrule
    \end{tabular}
    \vspace{\tablecapmargin}
    \caption{
        \textbf{Detection results using all available labels.}
        Baseline runtimes and model sizes are obtained from \cite{RVT}.
        For HMNet, we use the best-performing L3 variant.
        $^*$ Its runtime is computed using a V100 GPU, while T4 GPUs which are slower than V100 are used in the other cases.
        ERGO is {\color{gray}grayed} as it is the only method using pre-trained models (Swin Transformer V2~\cite{SwinTransformerV2}).
    }
    \label{tab:full-label-result}
    \vspace{\tablemargin}
    \vspace{-2mm}
\end{table}

\subsection{Ablation Studies}\label{sec:ablation}

% ablating each component
\begin{figure*}[t]
    \vspace{\pagetopmargin}
    \vspace{-5mm}
    \centering
    \begin{subfigure}{0.475\linewidth}
        \includegraphics[width=1.0\linewidth]{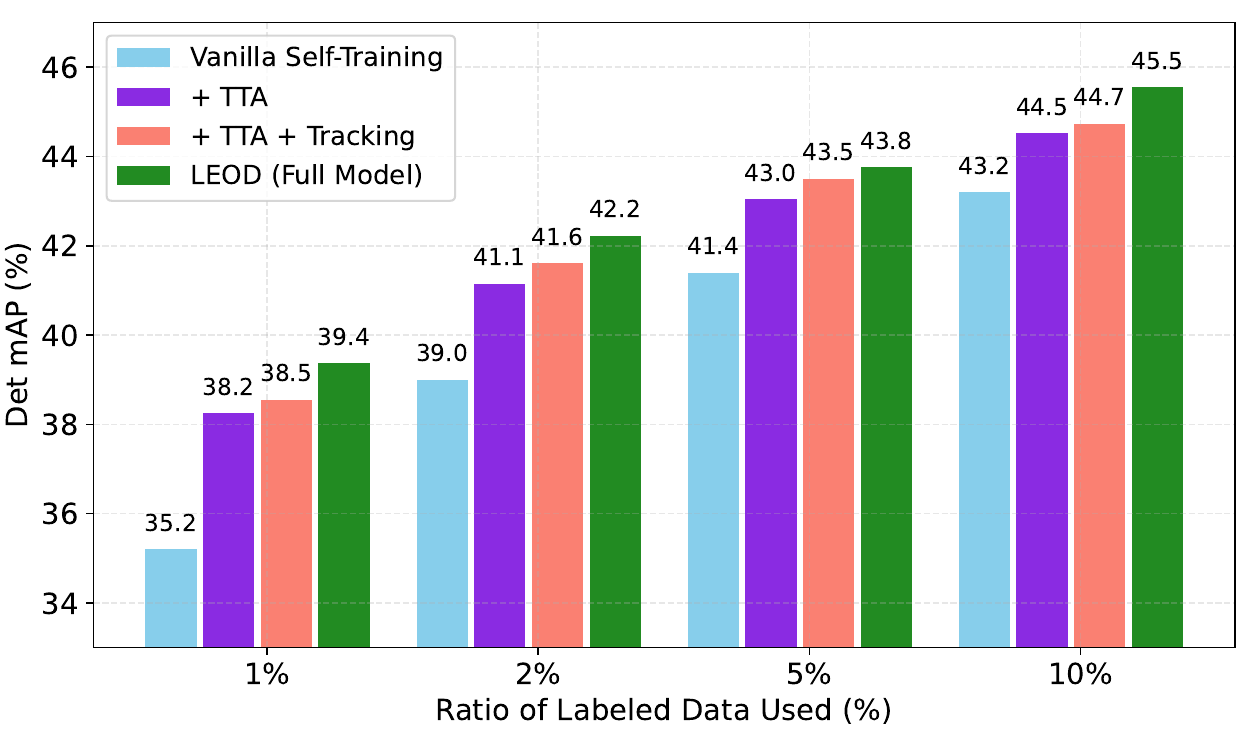}
        \vspace{-5.5mm}
        \caption{Gen1 weakly-supervised object detection (WSOD) results}
    \end{subfigure}
    \hspace{1mm}
    \begin{subfigure}{0.475\linewidth}
        \includegraphics[width=1.0\linewidth]{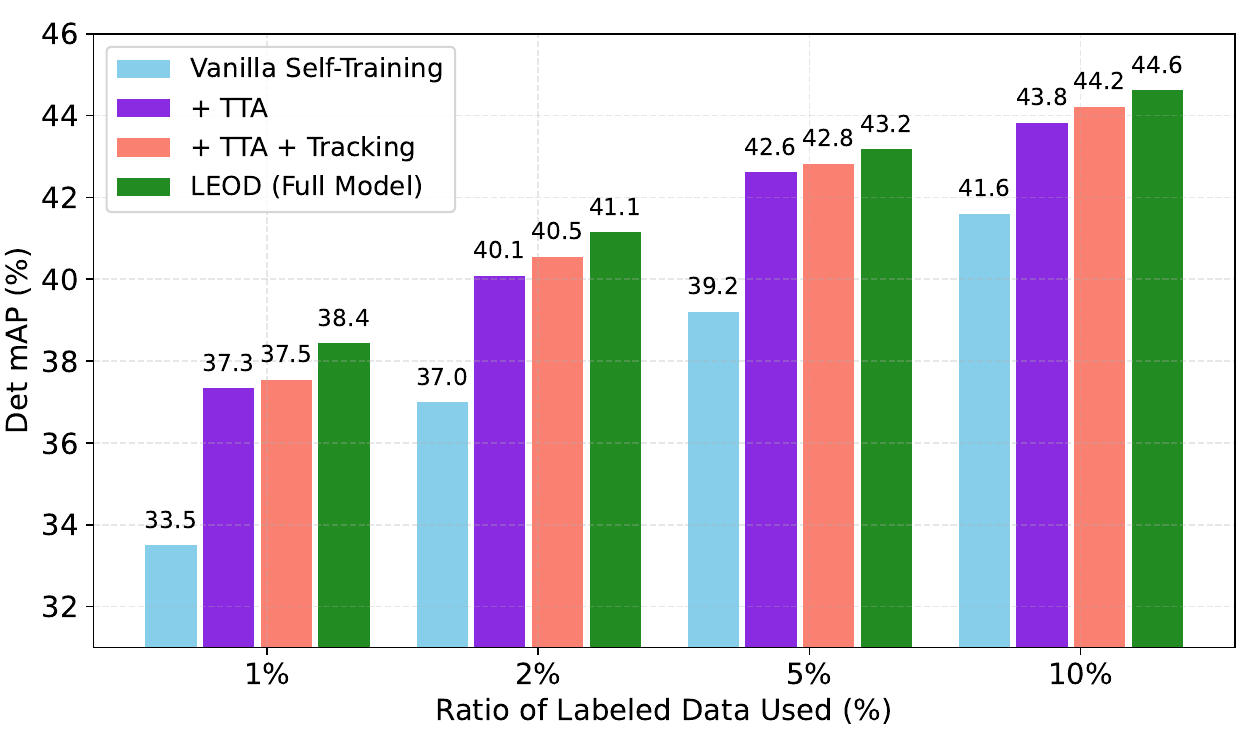}
        \vspace{-5.5mm}
        \caption{1Mpx weakly-supervised object detection (WSOD) results}
    \end{subfigure}
    \vspace{-2.5mm}
    \caption{
        \textbf{Ablation study of each component in \algoNameFull}.
        We report the WSOD result with RVT-S as the base detector on both datasets.
        Starting from Vanilla Self-Training, we gradually add TTA, tracking-based post-processing, and soft anchor assignment.
    }
    \label{fig:ablate-components}
    \vspace{\figmargin}
\end{figure*}

\heading{Larger Base Model.}
We evaluate our label-efficient learning framework on a stronger event-based detector RVT-B, which uses a larger ViT backbone compared to RVT-S.
\cref{tab:rvt-b-result} presents the result in the Gen1 WSOD setting.
With more parameters, RVT-B pre-trained on limited labels already outperforms RVT-S.
Still, \algoNameFull is able to improve the detection result by a sizeable margin across all labeling ratios.
With 100\% labels, our method achieves an mAP of 50.2\%, which is competitive with ERGO-12 using a large-scale pre-trained Swin Transformer V2 backbone~\cite{SwinTransformerV2}.
Notably, \algoNameFull brings larger absolute improvements on RVT-B compared to RVT-S, proving that our framework steadily scales up to enhance larger and stronger detectors.

% RVT-B
\begin{table}[t]
    \vspace{0.5mm}
    \centering
    \scriptsize
    \begin{tabular}{lccccc}
        \toprule
        \textbf{Method} & 1\% & 2\% & 5\% & 10\% & 100\% \\
        \midrule
        RVT-S & 30.8 & 34.4 & 37.9 & 40.8 & 46.5 \\
        \algoNameFull-RVT-S & 39.4 & 42.2 & 43.8 & 45.6 & 48.7 \\
        \midrule
        Absolute Improvement & +8.6 & +7.8 & +5.9 & +4.8 & +2.2 \\
        \specialrule{1pt}{2pt}{2pt}
        RVT-B & 31.6 & 34.8 & 38.3 & 41.0 & 47.6$^*$ \\
        \textbf{\algoNameFull-RVT-B} & \textbf{40.0} & \textbf{42.9} & \textbf{45.3} & \textbf{46.6} & \textbf{50.2} \\
        \midrule
        Absolute Improvement & +8.4 & +8.1 & +7.0 & +5.6 & +2.6 \\
        \bottomrule
    \end{tabular}
    \vspace{\tablecapmargin}
    \caption{
        \textbf{Gen1 WSOD mAPs (\%) with two RVT variants.}
        $^*$ our reproduced RVT-B using 100\% labels result is better than \cite{RVT}.
    }
    \label{tab:rvt-b-result}
    \vspace{\tablemargin}
\end{table}

% Number of self-training rounds
\begin{table}[t]
    \vspace{1mm}
    \centering
    \setlength{\tabcolsep}{3pt}
    \scriptsize
    \rowcolors{2}{uoftgray!50}{white}
    \begin{tabular}{cccccc|cccc}
        \toprule
        \textbf{Rounds} & 1\% & 2\% & 5\% & 10\% & 100\% & P. (1\%) & P. (2\%) & P. (5\%) & P. (10\%) \\
        \midrule
        1 & 38.1 & 41.1 & 43.1 & 45.3 & 48.5 & 0.65 & 0.69 & 0.74 & 0.79 \\
        2 & 39.4 & 42.2 & \textbf{43.8} & \textbf{45.6} & \textbf{48.7} & 0.72 & \textbf{0.75} & \textbf{0.77} & \textbf{0.81} \\
        3 & \textbf{39.5} & 42.2 & 43.6 & 45.4 & 48.6 & 0.72 & 0.74 & 0.74 & 0.76 \\
        \bottomrule
    \end{tabular}
    \vspace{\tablecapmargin}
    \caption{
        \textbf{Number of self-training rounds used in \algoNameFull.}
        We report the mAP (\%) result in Gen1 WSOD using RVT-S.
        We also compute the precision (P.) of pseudo labels that are used to train the models on that row, which is a good indicator for stop training.
    }
    \label{tab:num-training-round}
    \vspace{\tablemargin}
    \vspace{-1.5mm}
\end{table}

\heading{Effect of Each Component.}
\cref{fig:ablate-components} shows the model performance with different components on Gen1 and 1Mpx.
TTA significantly improves the mAP as it increases the recall in the generated pseudo labels using future information.
Leveraging the temporal persistency of objects, tracking-based post-processing leads to a further gain.
Finally, with soft anchor assignment, only reliable foreground and background labels are selected, easing the model training.

\heading{Self-Training Rounds.}
\cref{tab:num-training-round} left presents the results of multi-round self-training.
Detectors after the first round of self-training are significantly better than pre-trained models, and thus generate higher quality pseudo labels.
Therefore, a second round of training leads to consistent gains.
However, due to error accumulations, a third round of training may result in worse models.
To determine when to stop training, we empirically find that the precision of pseudo labels serves as a good indicator.
In \cref{tab:num-training-round} right, we compute the precision of predicted boxes on labeled frames (skipped labels are not used to prevent information leakage).
The precision consistently improves after round 1, but starts to decrease in some cases after round 2.
Indeed, the mAP also drops after training on those labels with lower precision.

% Hard/Soft thresholds
\begin{table}[t]
    \vspace{0.5mm}
    \centering
    \footnotesize
    \rowcolors{2}{uoftgray!50}{white}
    \begin{tabular}{lccccc}
        \toprule
        $(\tau_\text{hard, car}, \tau_\text{soft, car}, \tau_\text{soft, ped})$ & 1\% & 2\% & 5\% & 10\% \\
        \midrule
        $(0.5, 0.6, 0.3)$ & 38.9 & 41.8 & 43.5 & 45.4 \\
        $(\bm{0.6, 0.7, 0.35})$ & \textbf{39.4} & 42.2 & \textbf{43.8} & 45.5 \\
        $(0.6, 0.7, 0.4)$ & 39.3 & 42.0 & 43.7 & \textbf{45.6} \\
        $(0.6, 0.8, 0.4)$ & 39.3 & \textbf{42.3} & 43.6 & 45.5 \\
        $(0.7, 0.8, 0.4)$ & 39.0 & 42.0 & 43.2 & 44.9 \\
        $(0.8, -, -)^*$ & 38.4 & 41.4 & 42.7 & 44.2 \\
        \bottomrule
    \end{tabular}
    \vspace{\tablecapmargin}
    \caption{
        \textbf{Ablation of hard and soft thresholds used in soft anchor assignment.}
        We keep $\tau_\text{hard, ped} = \tau_\text{hard, car} / 2$.
        Ped stands for pedestrian.
        We report the mAP (\%) result in Gen1 WSOD using RVT-S.
        $^*$ indicates not using soft anchor assignment.
    }
    \label{tab:ablate-threshs}
    \vspace{\tablemargin}
    \vspace{-1mm}
\end{table}

\heading{Confidence Thresholds.}
We analyze the effects of hard and soft thresholds $(\tau_\text{hard}, \tau_\text{soft})$ in \cref{tab:ablate-threshs}.
Prior works~\cite{SoftTeacher,3DIoUMatchDet} often use a high threshold of 0.9, while we observe clearly lower mAPs when $\tau_\text{hard} > 0.7$.
With a lower $\tau_\text{hard}$, we retain most of the detected objects, and suppress noisy labels with $\tau_\text{soft}$.
\algoNameFull achieves similar results using several sets of hyper-parameters, showing our robustness to them.
See Appendix~\ref{app:pseudo-label-pr} for more analysis on the filtering thresholds.

\section{Conclusion}

We present \algoNameFull, the first algorithm for label-efficient event-based object detection.
To leverage unlabeled data, we adopt the self-training framework with reliable label selection.
Several techniques are introduced to improve labeling quality.
Extensive experiments on Gen1 and 1Mpx datasets showcase the superiority of our method over baselines in all the settings.

% CVPR
\heading{Limitations and Future Work.}
Following common practice, we only conduct intra-dataset experiments, \ie training on data gathered using the same protocol.
Recent works~\cite{SegmentAnything,RT-X,GPT-3} have shown that training large models over multiple datasets leads to excellent performance and generalization.
Since \algoNameFull benefits from unlabeled data, we can also train it jointly on more datasets that involve real-world multi-object event sequences~\cite{MVSEC,DSEC,M3ED}.
We discuss some failure cases of our pseudo-labeling method in Appendix~\ref{app:pseudo-label-vis}.

\subsection*{Acknowledgments}

This research was partially supported by Compute Ontario (\url{www.computeontario.ca}), the Province of Ontario, the Digital Research Alliance of Canada (\url{https://alliancecan.ca/}), and the Government of Canada through CIFAR and the Natural Sciences and Engineering Research Council of Canada (NSERC).
We would also like to acknowledge Vector Institute for computation support. We would like to thank Yash Kant, Lunjun Zhang, Xuanchi Ren, and Xiaoshi Wu for valuable discussions and support.

{
    \small
    \bibliographystyle{ieeenat_fullname}
    \bibliography{references}

\begin{thebibliography}{80}
\providecommand{\natexlab}[1]{#1}
\providecommand{\url}[1]{\texttt{#1}}
\expandafter\ifx\csname urlstyle\endcsname\relax
  \providecommand{\doi}[1]{doi: #1}\else
  \providecommand{\doi}{doi: \begingroup \urlstyle{rm}\Url}\fi

\bibitem[Araslanov and Roth(2020)]{WeaklySupSeg1}
Nikita Araslanov and Stefan Roth.
\newblock Single-stage semantic segmentation from image labels.
\newblock In \emph{CVPR}, 2020.

\bibitem[Bewley et~al.(2016)Bewley, Ge, Ott, Ramos, and Upcroft]{SORT}
Alex Bewley, Zongyuan Ge, Lionel Ott, Fabio Ramos, and Ben Upcroft.
\newblock Simple online and realtime tracking.
\newblock In \emph{ICIP}, 2016.

\bibitem[Brown et~al.(2020)Brown, Mann, Ryder, Subbiah, Kaplan, Dhariwal, Neelakantan, Shyam, Sastry, Askell, et~al.]{GPT-3}
Tom Brown, Benjamin Mann, Nick Ryder, Melanie Subbiah, Jared~D Kaplan, Prafulla Dhariwal, Arvind Neelakantan, Pranav Shyam, Girish Sastry, Amanda Askell, et~al.
\newblock Language models are few-shot learners.
\newblock \emph{NeurIPS}, 2020.

\bibitem[Chaney et~al.(2023)Chaney, Cladera, Wang, Bisulco, Hsieh, Korpela, Kumar, Taylor, and Daniilidis]{M3ED}
Kenneth Chaney, Fernando Cladera, Ziyun Wang, Anthony Bisulco, M.~Ani Hsieh, Christopher Korpela, Vijay Kumar, Camillo~J. Taylor, and Kostas Daniilidis.
\newblock M3ed: Multi-robot, multi-sensor, multi-environment event dataset.
\newblock In \emph{CVPRW}, 2023.

\bibitem[Chen(2018)]{EVCNNDet1}
Nicholas~FY Chen.
\newblock Pseudo-labels for supervised learning on dynamic vision sensor data, applied to object detection under ego-motion.
\newblock In \emph{CVPRW}, 2018.

\bibitem[Cho et~al.(2023)Cho, Kim, Chae, and Yoon]{Ev-LaFOR}
Hoonhee Cho, Hyeonseong Kim, Yujeong Chae, and Kuk-Jin Yoon.
\newblock Label-free event-based object recognition via joint learning with image reconstruction from events.
\newblock In \emph{ICCV}, 2023.

\bibitem[Cordone et~al.(2022)Cordone, Miramond, and Thierion]{SNNEvDet1}
Lo{\"\i}c Cordone, Beno{\^\i}t Miramond, and Philippe Thierion.
\newblock Object detection with spiking neural networks on automotive event data.
\newblock In \emph{IJCNN}, 2022.

\bibitem[De~Tournemire et~al.(2020)De~Tournemire, Nitti, Perot, Migliore, and Sironi]{Gen1Det}
Pierre De~Tournemire, Davide Nitti, Etienne Perot, Davide Migliore, and Amos Sironi.
\newblock A large scale event-based detection dataset for automotive.
\newblock \emph{arXiv preprint arXiv:2001.08499}, 2020.

\bibitem[Dosovitskiy et~al.(2021)Dosovitskiy, Beyer, Kolesnikov, Weissenborn, Zhai, Unterthiner, Dehghani, Minderer, Heigold, Gelly, et~al.]{ViT}
Alexey Dosovitskiy, Lucas Beyer, Alexander Kolesnikov, Dirk Weissenborn, Xiaohua Zhai, Thomas Unterthiner, Mostafa Dehghani, Matthias Minderer, Georg Heigold, Sylvain Gelly, et~al.
\newblock An image is worth 16x16 words: Transformers for image recognition at scale.
\newblock In \emph{ICLR}, 2021.

\bibitem[Finateu et~al.(2020)Finateu, Niwa, Matolin, Tsuchimoto, Mascheroni, Reynaud, Mostafalu, Brady, Chotard, LeGoff, et~al.]{Gen4_camera}
Thomas Finateu, Atsumi Niwa, Daniel Matolin, Koya Tsuchimoto, Andrea Mascheroni, Etienne Reynaud, Pooria Mostafalu, Frederick Brady, Ludovic Chotard, Florian LeGoff, et~al.
\newblock 5.10 a 1280$\times$ 720 back-illuminated stacked temporal contrast event-based vision sensor with 4.86 $\mu$m pixels, 1.066 geps readout, programmable event-rate controller and compressive data-formatting pipeline.
\newblock In \emph{2020 IEEE International Solid-State Circuits Conference}, 2020.

\bibitem[Gallego et~al.(2020)Gallego, Delbr{\"u}ck, Orchard, Bartolozzi, Taba, Censi, Leutenegger, Davison, Conradt, Daniilidis, et~al.]{EventVisionSurvey}
Guillermo Gallego, Tobi Delbr{\"u}ck, Garrick Orchard, Chiara Bartolozzi, Brian Taba, Andrea Censi, Stefan Leutenegger, Andrew~J Davison, J{\"o}rg Conradt, Kostas Daniilidis, et~al.
\newblock Event-based vision: A survey.
\newblock \emph{TPAMI}, 2020.

\bibitem[Ge et~al.(2021)Ge, Liu, Wang, Li, and Sun]{YOLOX}
Zheng Ge, Songtao Liu, Feng Wang, Zeming Li, and Jian Sun.
\newblock Yolox: Exceeding yolo series in 2021.
\newblock \emph{arXiv preprint arXiv:2107.08430}, 2021.

\bibitem[Gehrig and Scaramuzza(2022)]{EAGR}
Daniel Gehrig and Davide Scaramuzza.
\newblock Pushing the limits of asynchronous graph-based object detection with event cameras.
\newblock \emph{arXiv preprint arXiv:2211.12324}, 2022.

\bibitem[Gehrig and Scaramuzza(2023)]{RVT}
Mathias Gehrig and Davide Scaramuzza.
\newblock Recurrent vision transformers for object detection with event cameras.
\newblock In \emph{CVPR}, 2023.

\bibitem[Gehrig et~al.(2021)Gehrig, Aarents, Gehrig, and Scaramuzza]{DSEC}
Mathias Gehrig, Willem Aarents, Daniel Gehrig, and Davide Scaramuzza.
\newblock Dsec: A stereo event camera dataset for driving scenarios.
\newblock \emph{RA-L}, 2021.

\bibitem[Hamaguchi et~al.(2023)Hamaguchi, Furukawa, Onishi, and Sakurada]{HMNet}
Ryuhei Hamaguchi, Yasutaka Furukawa, Masaki Onishi, and Ken Sakurada.
\newblock Hierarchical neural memory network for low latency event processing.
\newblock In \emph{CVPR}, 2023.

\bibitem[Hochreiter and Schmidhuber(1997)]{LSTM}
Sepp Hochreiter and J{\"u}rgen Schmidhuber.
\newblock Long short-term memory.
\newblock \emph{Neural Computation}, 1997.

\bibitem[Hoffer et~al.(2017)Hoffer, Hubara, and Soudry]{SquareRootLR}
Elad Hoffer, Itay Hubara, and Daniel Soudry.
\newblock Train longer, generalize better: closing the generalization gap in large batch training of neural networks.
\newblock \emph{NeurIPS}, 2017.

\bibitem[Hu et~al.(2020)Hu, Delbruck, and Liu]{EventGraftNet}
Yuhuang Hu, Tobi Delbruck, and Shih-Chii Liu.
\newblock Learning to exploit multiple vision modalities by using grafted networks.
\newblock In \emph{ECCV}, 2020.

\bibitem[Hu et~al.(2021)Hu, Liu, and Delbruck]{V2E}
Yuhuang Hu, Shih-Chii Liu, and Tobi Delbruck.
\newblock v2e: From video frames to realistic dvs events.
\newblock In \emph{CVPR}, 2021.

\bibitem[Iacono et~al.(2018)Iacono, Weber, Glover, and Bartolozzi]{EVCNNDet2}
Massimiliano Iacono, Stefan Weber, Arren Glover, and Chiara Bartolozzi.
\newblock Towards event-driven object detection with off-the-shelf deep learning.
\newblock In \emph{IROS}, 2018.

\bibitem[Ji et~al.(2023)Ji, Wang, Yan, Liu, Xu, and Tang]{SNNEvDet3}
Mingcheng Ji, Ziling Wang, Rui Yan, Qingjie Liu, Shu Xu, and Huajin Tang.
\newblock Sctn: Event-based object tracking with energy-efficient deep convolutional spiking neural networks.
\newblock \emph{Frontiers in Neuroscience}, 2023.

\bibitem[Jiang et~al.(2019)Jiang, Xia, Huang, Stechele, Chen, Bing, and Knoll]{EVCNNDet3}
Zhuangyi Jiang, Pengfei Xia, Kai Huang, Walter Stechele, Guang Chen, Zhenshan Bing, and Alois Knoll.
\newblock Mixed frame-/event-driven fast pedestrian detection.
\newblock In \emph{ICRA}, 2019.

\bibitem[Khoreva et~al.(2017)Khoreva, Benenson, Hosang, Hein, and Schiele]{WeaklySupSeg3}
Anna Khoreva, Rodrigo Benenson, Jan Hosang, Matthias Hein, and Bernt Schiele.
\newblock Simple does it: Weakly supervised instance and semantic segmentation.
\newblock In \emph{CVPR}, 2017.

\bibitem[Kingma and Ba(2014)]{Adam}
Diederik~P Kingma and Jimmy Ba.
\newblock Adam: A method for stochastic optimization.
\newblock \emph{arXiv preprint arXiv:1412.6980}, 2014.

\bibitem[Kirillov et~al.(2023)Kirillov, Mintun, Ravi, Mao, Rolland, Gustafson, Xiao, Whitehead, Berg, Lo, et~al.]{SegmentAnything}
Alexander Kirillov, Eric Mintun, Nikhila Ravi, Hanzi Mao, Chloe Rolland, Laura Gustafson, Tete Xiao, Spencer Whitehead, Alexander~C Berg, Wan-Yen Lo, et~al.
\newblock Segment anything.
\newblock \emph{arXiv preprint arXiv:2304.02643}, 2023.

\bibitem[Klenk et~al.(2022)Klenk, Bonello, Koestler, and Cremers]{MEM}
Simon Klenk, David Bonello, Lukas Koestler, and Daniel Cremers.
\newblock Masked event modeling: Self-supervised pretraining for event cameras.
\newblock \emph{arXiv preprint arXiv:2212.10368}, 2022.

\bibitem[Lee et~al.(2013)]{SSCls1}
Dong-Hyun Lee et~al.
\newblock Pseudo-label: The simple and efficient semi-supervised learning method for deep neural networks.
\newblock In \emph{ICMLW}, 2013.

\bibitem[Li et~al.(2020)Li, Wu, Zhu, Xiong, Socher, and Davis]{DetwNoisyLabels}
Hengduo Li, Zuxuan Wu, Chen Zhu, Caiming Xiong, Richard Socher, and Larry~S Davis.
\newblock Learning from noisy anchors for one-stage object detection.
\newblock In \emph{CVPR}, 2020.

\bibitem[Li et~al.(2022)Li, Li, Zhu, Xiang, Huang, and Tian]{ASTMNet}
Jianing Li, Jia Li, Lin Zhu, Xijie Xiang, Tiejun Huang, and Yonghong Tian.
\newblock Asynchronous spatio-temporal memory network for continuous event-based object detection.
\newblock \emph{TIP}, 2022.

\bibitem[Lin et~al.(2017)Lin, Goyal, Girshick, He, and Doll{\'a}r]{Retina_Net}
Tsung-Yi Lin, Priya Goyal, Ross Girshick, Kaiming He, and Piotr Doll{\'a}r.
\newblock Focal loss for dense object detection.
\newblock In \emph{ICCV}, 2017.

\bibitem[Liu et~al.(2023)Liu, Xu, Yang, Yu, and Yu]{AED}
Bingde Liu, Chang Xu, Wen Yang, Huai Yu, and Lei Yu.
\newblock Motion robust high-speed light-weighted object detection with event camera.
\newblock \emph{IEEE Transactions on Instrumentation and Measurement}, 2023.

\bibitem[Liu et~al.(2022{\natexlab{a}})Liu, Zhou, Qi, Gong, Su, and Anguelov]{3DLiDARLESS}
Minghua Liu, Yin Zhou, Charles~R Qi, Boqing Gong, Hao Su, and Dragomir Anguelov.
\newblock Less: Label-efficient semantic segmentation for lidar point clouds.
\newblock In \emph{ECCV}, 2022{\natexlab{a}}.

\bibitem[Liu et~al.(2020)Liu, Ma, He, Kuo, Chen, Zhang, Wu, Kira, and Vajda]{UnbiasedTeacher}
Yen-Cheng Liu, Chih-Yao Ma, Zijian He, Chia-Wen Kuo, Kan Chen, Peizhao Zhang, Bichen Wu, Zsolt Kira, and Peter Vajda.
\newblock Unbiased teacher for semi-supervised object detection.
\newblock In \emph{ICLR}, 2020.

\bibitem[Liu et~al.(2022{\natexlab{b}})Liu, Ma, and Kira]{UnbiasedTeacherV2}
Yen-Cheng Liu, Chih-Yao Ma, and Zsolt Kira.
\newblock Unbiased teacher v2: Semi-supervised object detection for anchor-free and anchor-based detectors.
\newblock In \emph{CVPR}, 2022{\natexlab{b}}.

\bibitem[Liu et~al.(2022{\natexlab{c}})Liu, Hu, Lin, Yao, Xie, Wei, Ning, Cao, Zhang, Dong, et~al.]{SwinTransformerV2}
Ze Liu, Han Hu, Yutong Lin, Zhuliang Yao, Zhenda Xie, Yixuan Wei, Jia Ning, Yue Cao, Zheng Zhang, Li Dong, et~al.
\newblock Swin transformer v2: Scaling up capacity and resolution.
\newblock In \emph{CVPR}, 2022{\natexlab{c}}.

\bibitem[Messikommer et~al.(2020)Messikommer, Gehrig, Loquercio, and Scaramuzza]{EventSparseConv}
Nico Messikommer, Daniel Gehrig, Antonio Loquercio, and Davide Scaramuzza.
\newblock Event-based asynchronous sparse convolutional networks.
\newblock In \emph{ECCV}, 2020.

\bibitem[Messikommer et~al.(2022)Messikommer, Gehrig, Gehrig, and Scaramuzza]{EventDA}
Nico Messikommer, Daniel Gehrig, Mathias Gehrig, and Davide Scaramuzza.
\newblock Bridging the gap between events and frames through unsupervised domain adaptation.
\newblock \emph{RA-L}, 2022.

\bibitem[Misra et~al.(2015)Misra, Shrivastava, and Hebert]{SemiSupVidDet1}
Ishan Misra, Abhinav Shrivastava, and Martial Hebert.
\newblock Watch and learn: Semi-supervised learning for object detectors from video.
\newblock In \emph{CVPR}, 2015.

\bibitem[Mueggler et~al.(2017)Mueggler, Rebecq, Gallego, Delbruck, and Scaramuzza]{DAVIS-Sim}
Elias Mueggler, Henri Rebecq, Guillermo Gallego, Tobi Delbruck, and Davide Scaramuzza.
\newblock The event-camera dataset and simulator: Event-based data for pose estimation, visual odometry, and slam.
\newblock \emph{IJRR}, 2017.

\bibitem[Padalkar et~al.(2023)Padalkar, Pooley, Jain, Bewley, Herzog, Irpan, Khazatsky, Rai, Singh, Brohan, et~al.]{RT-X}
Abhishek Padalkar, Acorn Pooley, Ajinkya Jain, Alex Bewley, Alex Herzog, Alex Irpan, Alexander Khazatsky, Anant Rai, Anikait Singh, Anthony Brohan, et~al.
\newblock Open x-embodiment: Robotic learning datasets and rt-x models.
\newblock \emph{arXiv preprint arXiv:2310.08864}, 2023.

\bibitem[Papandreou et~al.(2015)Papandreou, Chen, Murphy, and Yuille]{WeaklySupSeg2}
George Papandreou, Liang-Chieh Chen, Kevin~P Murphy, and Alan~L Yuille.
\newblock Weakly-and semi-supervised learning of a deep convolutional network for semantic image segmentation.
\newblock In \emph{Proceedings of the IEEE international conference on computer vision}, pages 1742--1750, 2015.

\bibitem[Perot et~al.(2020)Perot, De~Tournemire, Nitti, Masci, and Sironi]{1MpxDet}
Etienne Perot, Pierre De~Tournemire, Davide Nitti, Jonathan Masci, and Amos Sironi.
\newblock Learning to detect objects with a 1 megapixel event camera.
\newblock \emph{NeurIPS}, 2020.

\bibitem[Posch et~al.(2010)Posch, Matolin, and Wohlgenannt]{Gen1_camera}
Christoph Posch, Daniel Matolin, and Rainer Wohlgenannt.
\newblock A qvga 143 db dynamic range frame-free pwm image sensor with lossless pixel-level video compression and time-domain cds.
\newblock \emph{IEEE Journal of Solid-State Circuits}, 2010.

\bibitem[Qi et~al.(2021)Qi, Zhou, Najibi, Sun, Vo, Deng, and Anguelov]{3DLiDARAutoLabeling}
Charles~R Qi, Yin Zhou, Mahyar Najibi, Pei Sun, Khoa Vo, Boyang Deng, and Dragomir Anguelov.
\newblock Offboard 3d object detection from point cloud sequences.
\newblock In \emph{CVPR}, 2021.

\bibitem[Rebecq et~al.(2018)Rebecq, Gehrig, and Scaramuzza]{ESIM}
Henri Rebecq, Daniel Gehrig, and Davide Scaramuzza.
\newblock Esim: an open event camera simulator.
\newblock In \emph{CoRL}, 2018.

\bibitem[Rebecq et~al.(2019{\natexlab{a}})Rebecq, Ranftl, Koltun, and Scaramuzza]{E2VID}
Henri Rebecq, Ren{\'e} Ranftl, Vladlen Koltun, and Davide Scaramuzza.
\newblock Events-to-video: Bringing modern computer vision to event cameras.
\newblock In \emph{CVPR}, 2019{\natexlab{a}}.

\bibitem[Rebecq et~al.(2019{\natexlab{b}})Rebecq, Ranftl, Koltun, and Scaramuzza]{E2VID-PAMI}
Henri Rebecq, Ren{\'e} Ranftl, Vladlen Koltun, and Davide Scaramuzza.
\newblock High speed and high dynamic range video with an event camera.
\newblock \emph{TPAMI}, 2019{\natexlab{b}}.

\bibitem[Ren et~al.(2015)Ren, He, Girshick, and Sun]{Faster_RCNN}
Shaoqing Ren, Kaiming He, Ross Girshick, and Jian Sun.
\newblock Faster r-cnn: Towards real-time object detection with region proposal networks.
\newblock \emph{NeurIPS}, 2015.

\bibitem[Schaefer et~al.(2022)Schaefer, Gehrig, and Scaramuzza]{AEGNN}
Simon Schaefer, Daniel Gehrig, and Davide Scaramuzza.
\newblock Aegnn: Asynchronous event-based graph neural networks.
\newblock In \emph{CVPR}, 2022.

\bibitem[Scheerlinck et~al.(2020)Scheerlinck, Rebecq, Gehrig, Barnes, Mahony, and Scaramuzza]{FireNet}
Cedric Scheerlinck, Henri Rebecq, Daniel Gehrig, Nick Barnes, Robert Mahony, and Davide Scaramuzza.
\newblock Fast image reconstruction with an event camera.
\newblock In \emph{WACV}, 2020.

\bibitem[Shi et~al.(2020)Shi, Guo, Jiang, Wang, Shi, Wang, and Li]{PV-RCNN}
Shaoshuai Shi, Chaoxu Guo, Li Jiang, Zhe Wang, Jianping Shi, Xiaogang Wang, and Hongsheng Li.
\newblock Pv-rcnn: Point-voxel feature set abstraction for 3d object detection.
\newblock In \emph{CVPR}, 2020.

\bibitem[Shi et~al.(2015)Shi, Chen, Wang, Yeung, Wong, and Woo]{ConvLSTM}
Xingjian Shi, Zhourong Chen, Hao Wang, Dit-Yan Yeung, Wai-Kin Wong, and Wang-chun Woo.
\newblock Convolutional lstm network: A machine learning approach for precipitation nowcasting.
\newblock \emph{NeurIPS}, 2015.

\bibitem[Singh et~al.(2016)Singh, Xiao, and Lee]{WeaklySupVidDet1}
Krishna~Kumar Singh, Fanyi Xiao, and Yong~Jae Lee.
\newblock Track and transfer: Watching videos to simulate strong human supervision for weakly-supervised object detection.
\newblock In \emph{Proceedings of the IEEE Conference on Computer Vision and Pattern Recognition}, pages 3548--3556, 2016.

\bibitem[Sohn et~al.(2020{\natexlab{a}})Sohn, Berthelot, Carlini, Zhang, Zhang, Raffel, Cubuk, Kurakin, and Li]{FixMatch}
Kihyuk Sohn, David Berthelot, Nicholas Carlini, Zizhao Zhang, Han Zhang, Colin~A Raffel, Ekin~Dogus Cubuk, Alexey Kurakin, and Chun-Liang Li.
\newblock Fixmatch: Simplifying semi-supervised learning with consistency and confidence.
\newblock \emph{NeurIPS}, 2020{\natexlab{a}}.

\bibitem[Sohn et~al.(2020{\natexlab{b}})Sohn, Zhang, Li, Zhang, Lee, and Pfister]{STAC}
Kihyuk Sohn, Zizhao Zhang, Chun-Liang Li, Han Zhang, Chen-Yu Lee, and Tomas Pfister.
\newblock A simple semi-supervised learning framework for object detection.
\newblock \emph{arXiv preprint arXiv:2005.04757}, 2020{\natexlab{b}}.

\bibitem[Stoffregen et~al.(2020)Stoffregen, Scheerlinck, Scaramuzza, Drummond, Barnes, Kleeman, and Mahony]{BetterE2VID}
Timo Stoffregen, Cedric Scheerlinck, Davide Scaramuzza, Tom Drummond, Nick Barnes, Lindsay Kleeman, and Robert Mahony.
\newblock Reducing the sim-to-real gap for event cameras.
\newblock In \emph{ECCV}, 2020.

\bibitem[Sun et~al.(2022)Sun, Messikommer, Gehrig, and Scaramuzza]{ESS_EvSegFromImg}
Zhaoning Sun, Nico Messikommer, Daniel Gehrig, and Davide Scaramuzza.
\newblock Ess: Learning event-based semantic segmentation from still images.
\newblock In \emph{ECCV}, 2022.

\bibitem[Tu et~al.(2022)Tu, Talebi, Zhang, Yang, Milanfar, Bovik, and Li]{MaxViT}
Zhengzhong Tu, Hossein Talebi, Han Zhang, Feng Yang, Peyman Milanfar, Alan Bovik, and Yinxiao Li.
\newblock Maxvit: Multi-axis vision transformer.
\newblock In \emph{ECCV}, 2022.

\bibitem[Van~Engelen and Hoos(2020)]{semi-sup-survey1}
Jesper~E Van~Engelen and Holger~H Hoos.
\newblock A survey on semi-supervised learning.
\newblock \emph{Machine learning}, 2020.

\bibitem[Wang et~al.(2021{\natexlab{a}})Wang, Cong, Litany, Gao, and Guibas]{3DIoUMatchDet}
He Wang, Yezhen Cong, Or Litany, Yue Gao, and Leonidas~J Guibas.
\newblock 3dioumatch: Leveraging iou prediction for semi-supervised 3d object detection.
\newblock In \emph{CVPR}, 2021{\natexlab{a}}.

\bibitem[Wang et~al.(2021{\natexlab{b}})Wang, Chae, Yoon, Kim, and Yoon]{Evdistill}
Lin Wang, Yujeong Chae, Sung-Hoon Yoon, Tae-Kyun Kim, and Kuk-Jin Yoon.
\newblock Evdistill: Asynchronous events to end-task learning via bidirectional reconstruction-guided cross-modal knowledge distillation.
\newblock In \emph{CVPR}, 2021{\natexlab{b}}.

\bibitem[Wang et~al.(2023)Wang, Girdhar, Yu, and Misra]{CutLer}
Xudong Wang, Rohit Girdhar, Stella~X Yu, and Ishan Misra.
\newblock Cut and learn for unsupervised object detection and instance segmentation.
\newblock In \emph{CVPR}, 2023.

\bibitem[Weng et~al.(2021)Weng, Zhang, and Xiong]{E2VID-Transformer}
Wenming Weng, Yueyi Zhang, and Zhiwei Xiong.
\newblock Event-based video reconstruction using transformer.
\newblock In \emph{ICCV}, 2021.

\bibitem[Wu et~al.(2023)Wu, Liu, and Gilitschenski]{EventCLIP}
Ziyi Wu, Xudong Liu, and Igor Gilitschenski.
\newblock Eventclip: Adapting clip for event-based object recognition.
\newblock \emph{arXiv preprint arXiv:2306.06354}, 2023.

\bibitem[Xie et~al.(2020)Xie, Luong, Hovy, and Le]{NoisyStudent}
Qizhe Xie, Minh-Thang Luong, Eduard Hovy, and Quoc~V Le.
\newblock Self-training with noisy student improves imagenet classification.
\newblock In \emph{CVPR}, 2020.

\bibitem[Xu et~al.(2021)Xu, Zhang, Hu, Wang, Wang, Wei, Bai, and Liu]{SoftTeacher}
Mengde Xu, Zheng Zhang, Han Hu, Jianfeng Wang, Lijuan Wang, Fangyun Wei, Xiang Bai, and Zicheng Liu.
\newblock End-to-end semi-supervised object detection with soft teacher.
\newblock In \emph{ICCV}, 2021.

\bibitem[Yan et~al.(2019)Yan, Li, Xie, Li, Wang, Chen, and Lin]{SemiSupVidDet2}
Pengxiang Yan, Guanbin Li, Yuan Xie, Zhen Li, Chuan Wang, Tianshui Chen, and Liang Lin.
\newblock Semi-supervised video salient object detection using pseudo-labels.
\newblock In \emph{ICCV}, 2019.

\bibitem[Yang et~al.(2022)Yang, Song, King, and Xu]{semi-sup-survey2}
Xiangli Yang, Zixing Song, Irwin King, and Zenglin Xu.
\newblock A survey on deep semi-supervised learning.
\newblock \emph{TKDE}, 2022.

\bibitem[Yang et~al.(2023)Yang, Pan, and Liu]{EvDataPretrain}
Yan Yang, Liyuan Pan, and Liu Liu.
\newblock Event camera data pre-training.
\newblock In \emph{ICCV}, 2023.

\bibitem[Yin et~al.(2022)Yin, Fang, Zhou, Zhang, Xu, Shen, and Wang]{3DDetProficientTeachers}
Junbo Yin, Jin Fang, Dingfu Zhou, Liangjun Zhang, Cheng-Zhong Xu, Jianbing Shen, and Wenguan Wang.
\newblock Semi-supervised 3d object detection with proficient teachers.
\newblock In \emph{ECCV}, 2022.

\bibitem[Zanardi et~al.(2019)Zanardi, Aumiller, Zilly, Censi, and Frazzoli]{WormholeLearningRSS}
Alessandro Zanardi, Andreas Aumiller, Julian Zilly, Andrea Censi, and Emilio Frazzoli.
\newblock Cross-modal learning filters for rgb-neuromorphic wormhole learning.
\newblock \emph{RSS}, 2019.

\bibitem[Zhang et~al.(2022)Zhang, Dong, Zhang, Ding, Heide, Yin, and Yang]{SNNEvDet2}
Jiqing Zhang, Bo Dong, Haiwei Zhang, Jianchuan Ding, Felix Heide, Baocai Yin, and Xin Yang.
\newblock Spiking transformers for event-based single object tracking.
\newblock In \emph{CVPR}, 2022.

\bibitem[Zhang et~al.(2023)Zhang, Yang, Xiong, Casas, Yang, Ren, and Urtasun]{Oyster}
Lunjun Zhang, Anqi~Joyce Yang, Yuwen Xiong, Sergio Casas, Bin Yang, Mengye Ren, and Raquel Urtasun.
\newblock Towards unsupervised object detection from lidar point clouds.
\newblock In \emph{CVPR}, 2023.

\bibitem[Zhang et~al.(2018)Zhang, Bai, Ding, Li, and Ghanem]{WeaklySup2DDet1}
Yongqiang Zhang, Yancheng Bai, Mingli Ding, Yongqiang Li, and Bernard Ghanem.
\newblock W2f: A weakly-supervised to fully-supervised framework for object detection.
\newblock In \emph{CVPR}, 2018.

\bibitem[Zhou(2018)]{weakly-sup-survey1}
Zhi-Hua Zhou.
\newblock A brief introduction to weakly supervised learning.
\newblock \emph{National Science Review}, 2018.

\bibitem[Zhu et~al.(2018)Zhu, Thakur, {\"O}zaslan, Pfrommer, Kumar, and Daniilidis]{MVSEC}
Alex~Zihao Zhu, Dinesh Thakur, Tolga {\"O}zaslan, Bernd Pfrommer, Vijay Kumar, and Kostas Daniilidis.
\newblock The multivehicle stereo event camera dataset: An event camera dataset for 3d perception.
\newblock \emph{RA-L}, 2018.

\bibitem[Zhu et~al.(2019)Zhu, Yuan, Chaney, and Daniilidis]{UnsupEvOptFlow}
Alex~Zihao Zhu, Liangzhe Yuan, Kenneth Chaney, and Kostas Daniilidis.
\newblock Unsupervised event-based learning of optical flow, depth, and egomotion.
\newblock In \emph{CVPR}, 2019.

\bibitem[Zhu et~al.(2021)Zhu, Wang, Khant, and Daniilidis]{EventGAN}
Alex~Zihao Zhu, Ziyun Wang, Kaung Khant, and Kostas Daniilidis.
\newblock Eventgan: Leveraging large scale image datasets for event cameras.
\newblock In \emph{ICCP}, 2021.

\bibitem[Zubi{\'c} et~al.(2023)Zubi{\'c}, Gehrig, Gehrig, and Scaramuzza]{GWDDet}
Nikola Zubi{\'c}, Daniel Gehrig, Mathias Gehrig, and Davide Scaramuzza.
\newblock From chaos comes order: Ordering event representations for object recognition and detection.
\newblock In \emph{ICCV}, 2023.

\end{thebibliography}
}

\clearpage

\appendix

\section{More Implementation Details}\label{app:more-implementation-details}

\subsection{Tracking-based Post-Processing}\label{app:tracking}

Given the detection outputs from TTA, we first aggregate them via Non-Maximum Suppression (NMS).
Now, for each event frame $I$ at timestep $t$, we have a set of 2D bounding boxes $\mathcal{B}^t = \{b^t_j = (x_j, y_j, w_j, h_j, l_j, t)\}$.
We follow the tracking-by-detection paradigm~\cite{SORT} to build tracks by linking detection boxes between frames, which is also inspired by \cite{Oyster}.
Each track $s_k = \{(b, v_x, v_y)_t, k, n, q\}$ maintains the following attributes: $(v_x, v_y)$ is the estimated velocity in the pixel space, $k$ is the track's unique ID, $n$ is its length so far, and $q \in [0, 1]$ is its current score, which is decayed over time and determines whether to delete the track.
In the first frame, we initialize each box in $\mathcal{B}^0$ as a track, where $(v_x, v_y) = (0, 0)$, $n = 1$, and $q = 0.9$.
For every coming frame $I_t$, we need to link its bounding boxes $\mathcal{B}^t$ to existing tracks $\{s_k\}$.
We first predict the new box parameter of each track using its coordinate in the last frame $(x, y)$ and $(v_x, v_y)$ with a linear motion assumption, while keeping its size in the last frame $(w, h)$ unchanged.
Then, we compute pairwise IoUs between the predicted boxes and $\mathcal{B}^t$ to apply greedy matching.
Only boxes in the same category and with an IoU larger than $\tau_\text{iou}$ can be matched.
For unmatched boxes, we initialize tracks for them as done in the first frame.
For unmatched tracks, we decay its score as $q_t = 0.9 * q_{t-1}$, which allows for object re-identification in future frames.
For matched boxes and tracks, we update the box parameters and velocity, and reset the score as $q = 1$.
Finally, we go over each track and delete those with a lower score $q < \tau_\text{del}$.
After tracking, each box is associated with a track, and thus a length $n$ (note that $n$ represents the number of successful matches instead of the time between creation and deletion, \ie unmatched timesteps do not count).
We identify boxes with $n < T_{trk}$ as temporally inconsistent.

Similar to TTA, we apply tracking in forward and backward event sequences, and will only remove a box if it has a short track length in both directions.
For those long tracks, we inpaint boxes at their unmatched timesteps using the synthesized ones with linear motion.
This builds upon the prior of object permanence and can further stabilize the training in our experiments.
Overall, the detection losses related to removed and inpainted boxes will be ignored during model training.
For hyper-parameters, we choose $\tau_\text{iou} = 0.45$ which is the same as the IoU threshold used in NMS, $\tau_\text{del} = 0.55$ which is slightly higher than $0.9^6 \approx 0.53$, and $T_{trk} = 6$.
We do not tune these hyper-parameters and simply use the first set of values that works.

\subsection{RVT Training}\label{app:rvt-training}

We build upon the open-source codebase of RVT\footnote{\url{https://github.com/uzh-rpg/RVT}}~\cite{RVT} and copy most of their training settings.
Events in each $50ms$ time window are converted to a frame-like 10-channel event histogram representation.
We use RVT-S in most of the experiments due to limited computation resources, but also scale up \algoNameFull to RVT-B in Sec. 4.4.
Following \cite{RVT}, we down-sample the labeling frequency of 1Mpx~\cite{1MpxDet} to 10 Hz.

\heading{Pre-training on Sparse Labels.}
The same optimizer, batch size, data augmentation, and data sampling methods are used.
In order to apply the time-flip TTA during pseudo-labeling, we add a temporal flipping augmentation.
We train for 200k steps on 1\% labels, 300k steps on 2\% labels, and 400k steps on 5\%, 10\%, and 100\% labels.
On 1Mpx~\cite{1MpxDet}, we use an increased sequence length $L = 10$ for training, as we observed clearly better results compared to $L = 5$.

\heading{Pseudo Label Filtering.}
We filter out low-confidence bounding boxes to obtain high-quality pseudo labels.
As introduced in Sec. 3.1, RVT predicts an objectness score $p_{obj} \in [0, 1]$ and a class-wise IoU score $p_{iou} \in \mathbb{R}^C, p_{iou}^i \in [0, 1]$ for each bounding box.
We only keep boxes with $p_{obj} \ge \tau_\text{hard}$ and $\text{max}(p_{iou}) \ge \tau_\text{hard}$, and further ignore losses on those with $p_{obj} < \tau_\text{soft}$ and $\text{max}(p_{iou}) < \tau_\text{soft}$.

\heading{Self-training on Pseudo Labels.}
We still use the same batch size, data augmentation, and data sampling methods.
Since pseudo labels have a much higher labeling frequency than the original ground-truth labels, the effective training batch size under the same event sequence length is larger.
Following the square root scaling law~\cite{SquareRootLR}, we use a higher learning rate of $5\times10^{-4}$ on Gen1~\cite{Gen1Det} and $8\times10^{-4}$ on 1Mpx.
We train for 150k and 200k steps in round 1 and round 2 self-training, respectively.
At each training step, we first conduct the normal anchor assignment process~\cite{YOLOX} to compute training losses, and then set the losses on anchors associated with uncertain boxes (boxes with a detection score lower than $\tau_\text{soft}$ and the ignored and inpainted boxes from tracking-based post-processing) as 0.

\heading{Training Objective of RVT.}
RVT adopts the anchor-free YOLOX~\cite{YOLOX} detection head.
Let $o^{i} \in \{0, 1\}$ denote whether an anchor point is matched to a ground-truth box kept after label filtering, and $r^{i} \in \{0, 1\}$ denote whether it is matched to a box removed in tracking or soft anchor assignment (thus ignored in loss computation), the training loss of RVT is:
\begin{equation}
\vspace{-1mm}
\begin{aligned}
  L = & \mathbbm{1}_{\{r^{i} = 0\}} L_\text{BCE}(p_\text{obj}^{i}, o^{i}) \ + \ \mathbbm{1}_{\{o^{i} = 1\}} L_\text{CE}(p_\text{iou}^{i}, l^{i}) \ \\
    &+ \ \mathbbm{1}_{\{o^{i} = 1\}} L_\text{IoU}(\Delta b^{i}, b^{i})
\end{aligned}
\end{equation}
\vspace{-1mm}

Our proposed components only bring negligible overheads to model training.
Therefore, we can train our model on 2 NVIDIA A40 GPUs.
The pre-training stage takes 60 hours, while self-training takes around 40 hours.

% comprehensive comparison of pseudo label Precision vs Recall
\begin{figure*}[t]
    \vspace{\pagetopmargin}
    \vspace{-2mm}
    \centering
    \begin{subfigure}{0.99\linewidth}
        \includegraphics[width=1.0\linewidth]{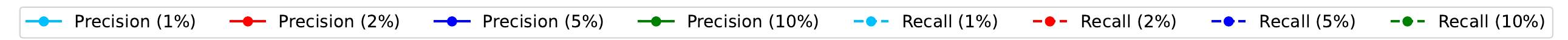}
    \end{subfigure}
    \\ \vspace{0mm}
    \begin{subfigure}{0.335\linewidth}
        \includegraphics[width=1.0\linewidth]{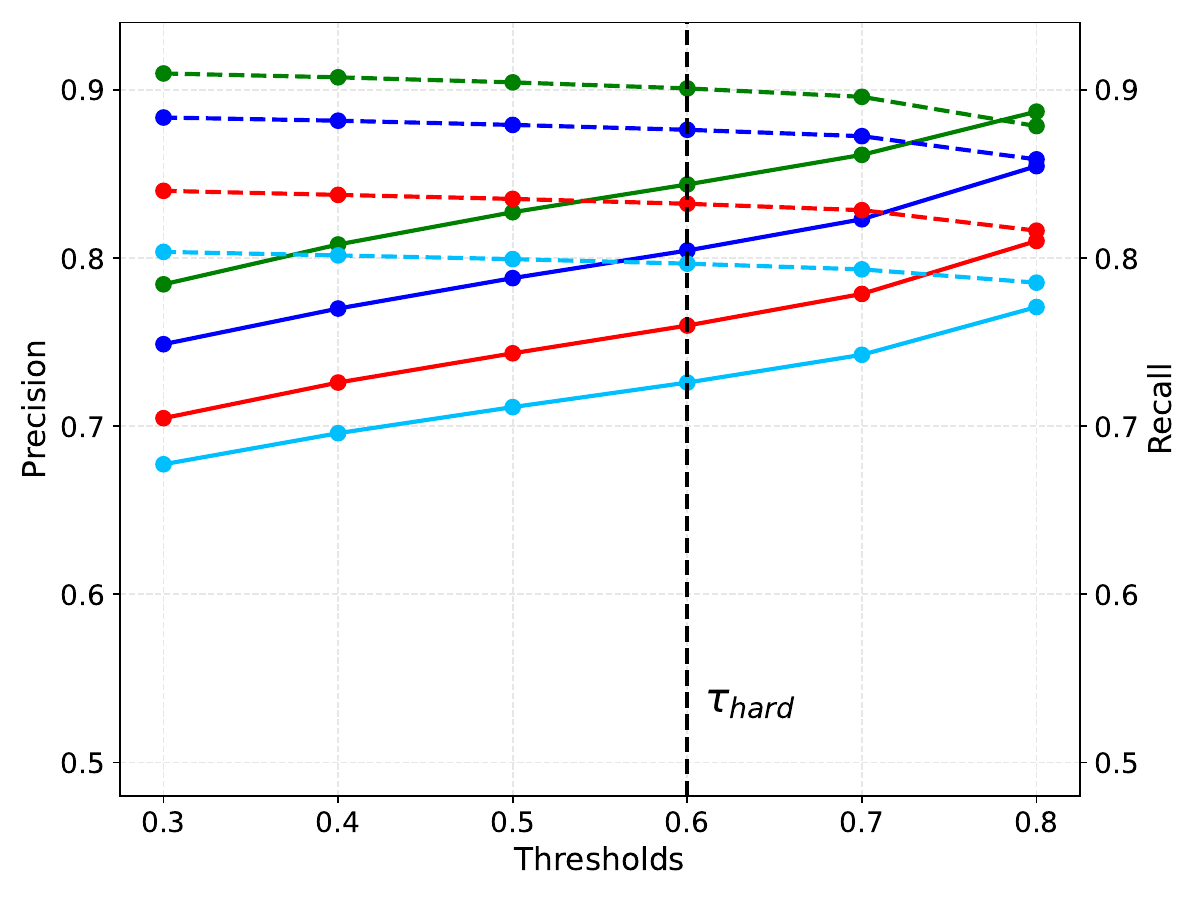}
        \vspace{-5mm}
        \caption{Gen1 WSOD Cars}
    \end{subfigure}
    \hspace{-2.5mm}
    \begin{subfigure}{0.335\linewidth}
        \includegraphics[width=1.0\linewidth]{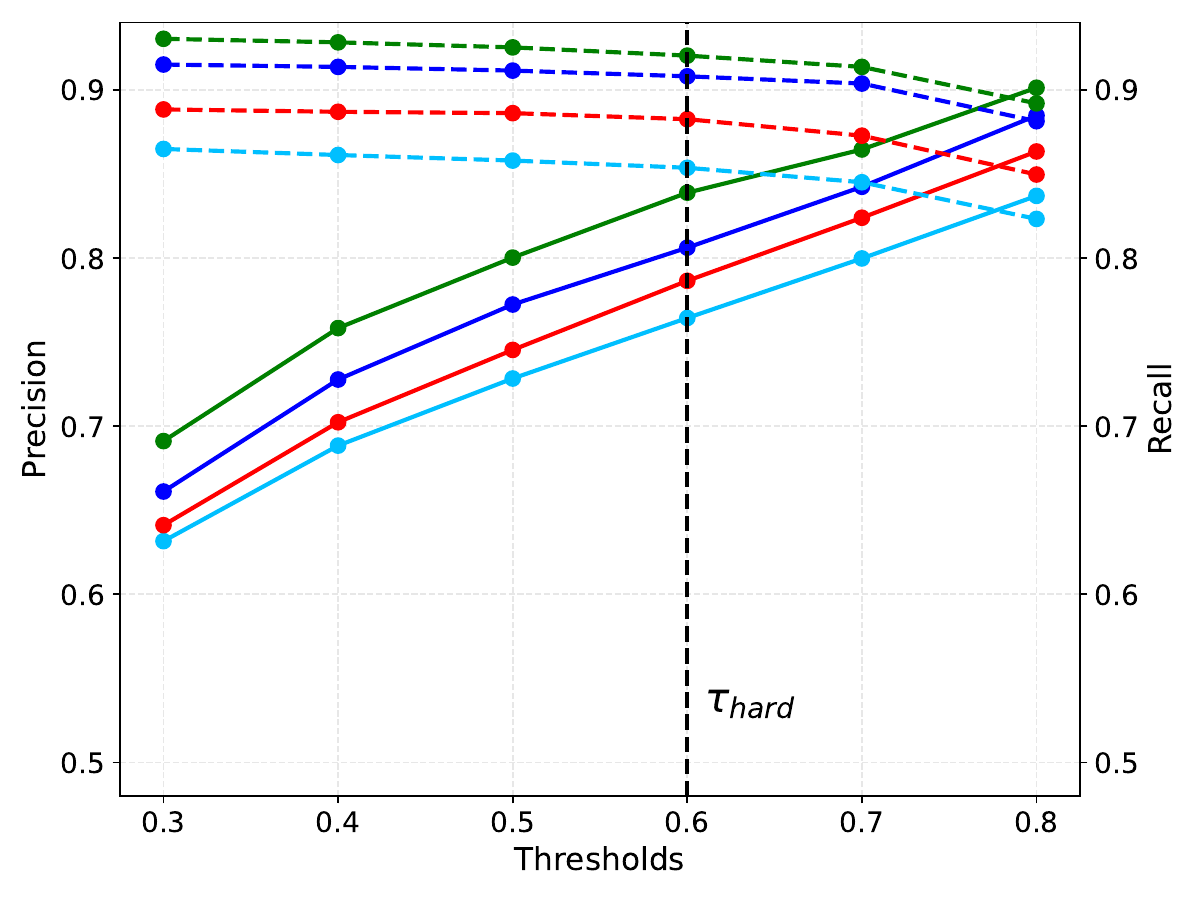}
        \vspace{-5mm}
        \caption{Gen1 WSOD 1Round Cars}
    \end{subfigure}
    \hspace{-2.5mm}
    \begin{subfigure}{0.335\linewidth}
        \includegraphics[width=1.0\linewidth]{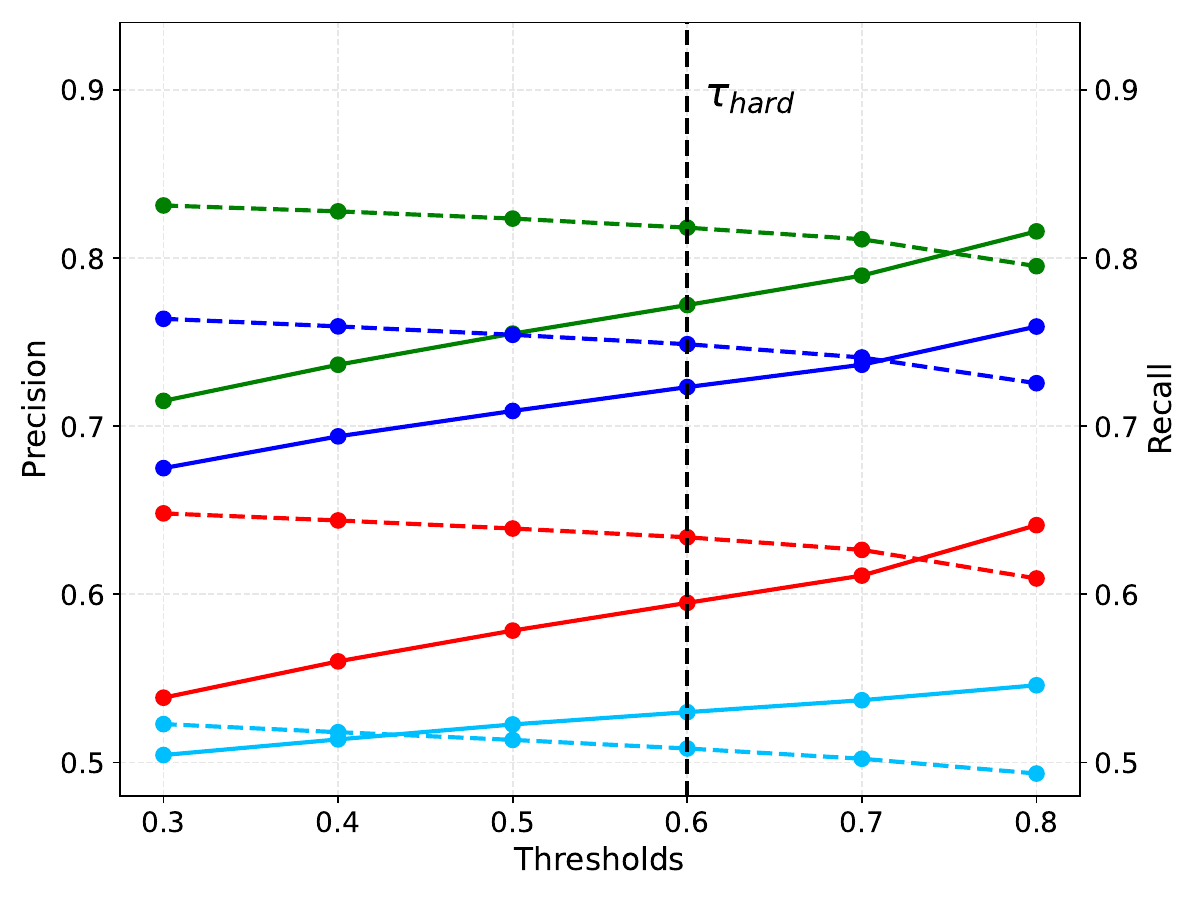}
        \vspace{-5mm}
        \caption{Gen1 SSOD Cars}
    \end{subfigure}
    \\ \vspace{1mm}
    \begin{subfigure}{0.335\linewidth}
        \includegraphics[width=1.0\linewidth]{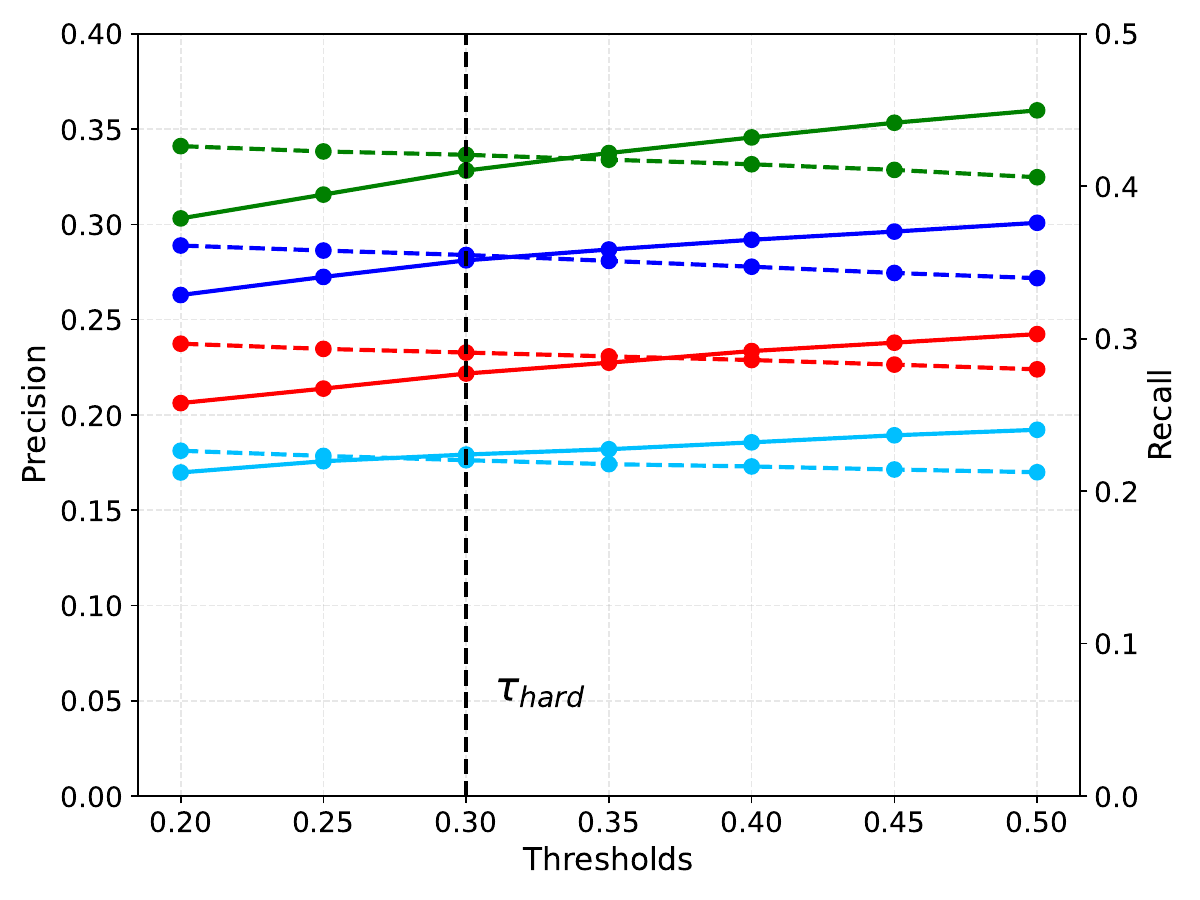}
        \vspace{-5mm}
        \caption{Gen1 WSOD Pedestrians}
    \end{subfigure}
    \hspace{-2.5mm}
    \begin{subfigure}{0.335\linewidth}
        \includegraphics[width=1.0\linewidth]{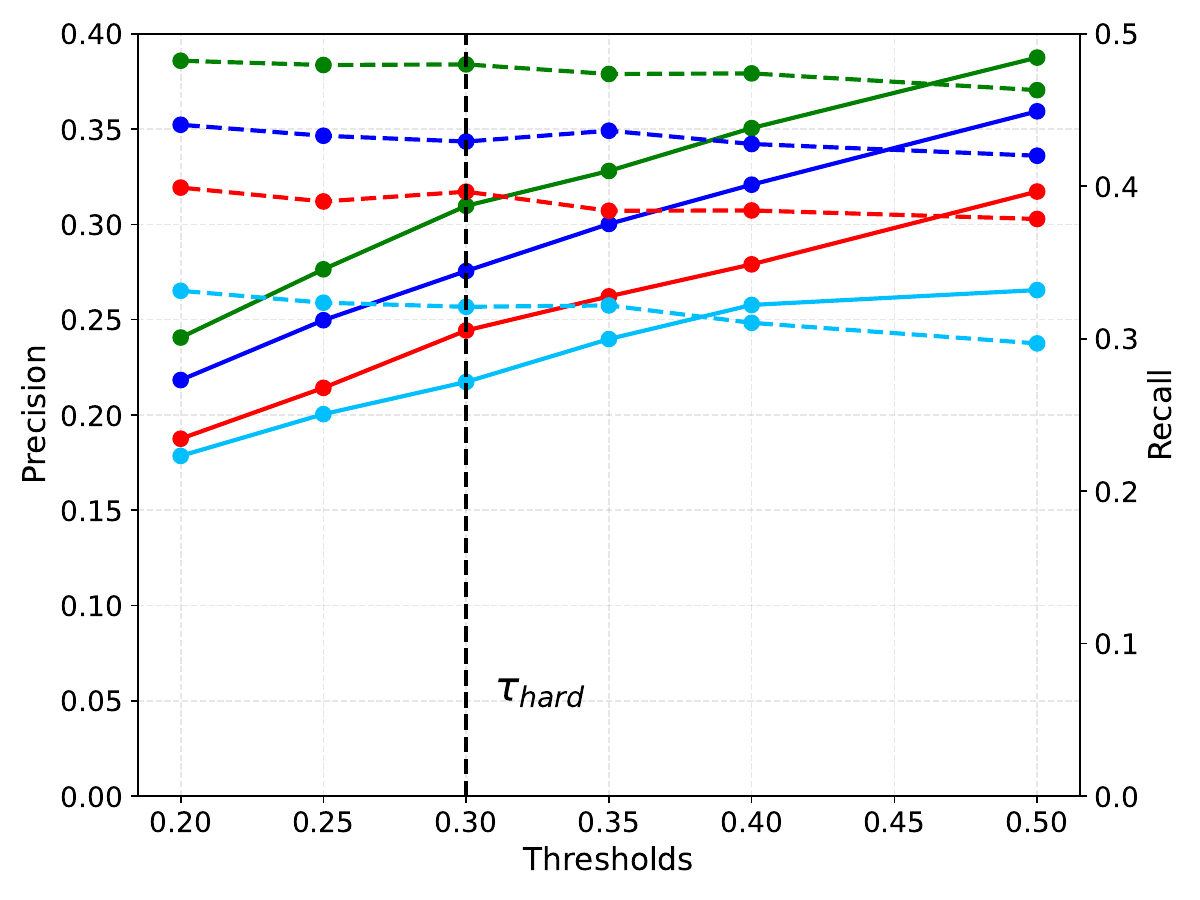}
        \vspace{-5mm}
        \caption{Gen1 WSOD 1Round Pedestrians}
    \end{subfigure}
    \hspace{-2.5mm}
    \begin{subfigure}{0.335\linewidth}
        \includegraphics[width=1.0\linewidth]{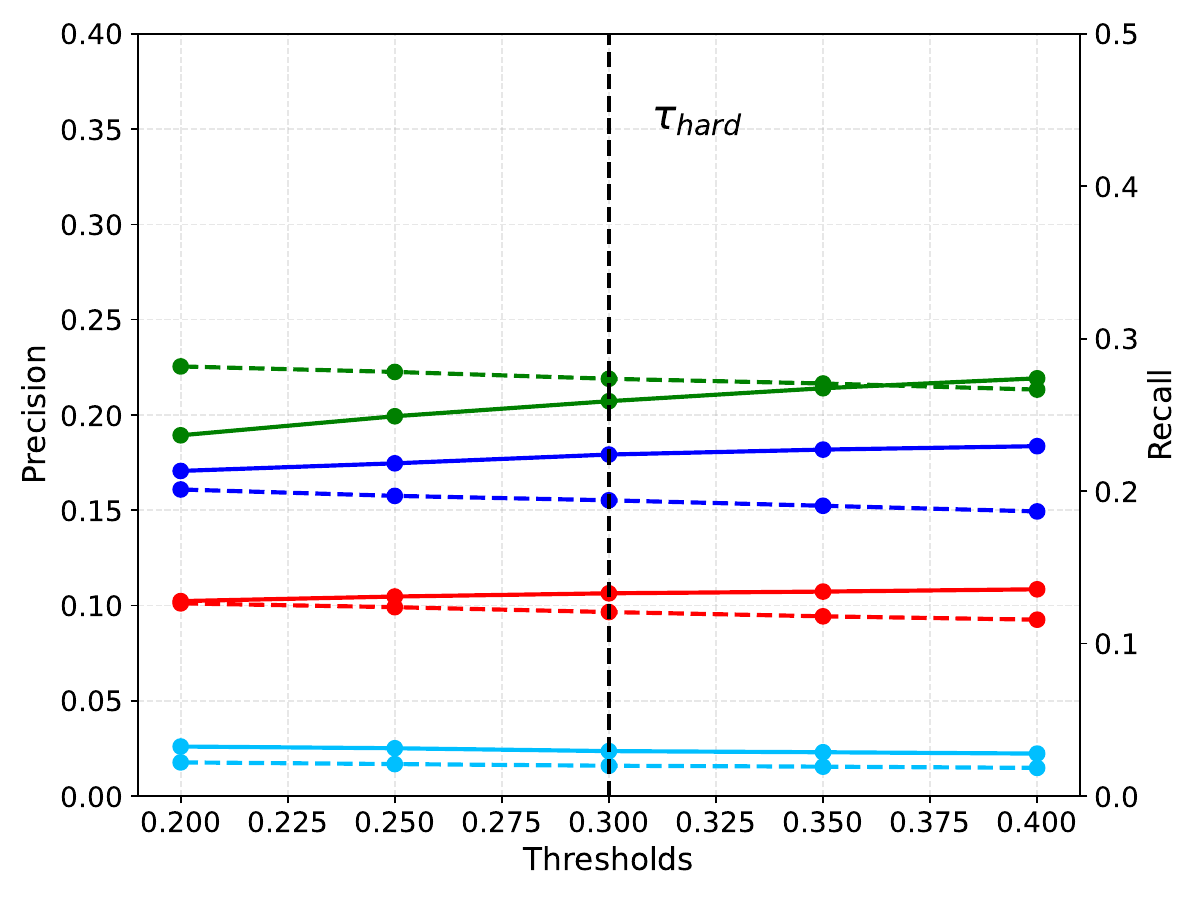}
        \vspace{-5mm}
        \caption{Gen1 SSOD Pedestrians}
    \end{subfigure}
    \\ \vspace{1mm}
    % 1Mpx
    \begin{subfigure}{0.335\linewidth}
        \includegraphics[width=1.0\linewidth]{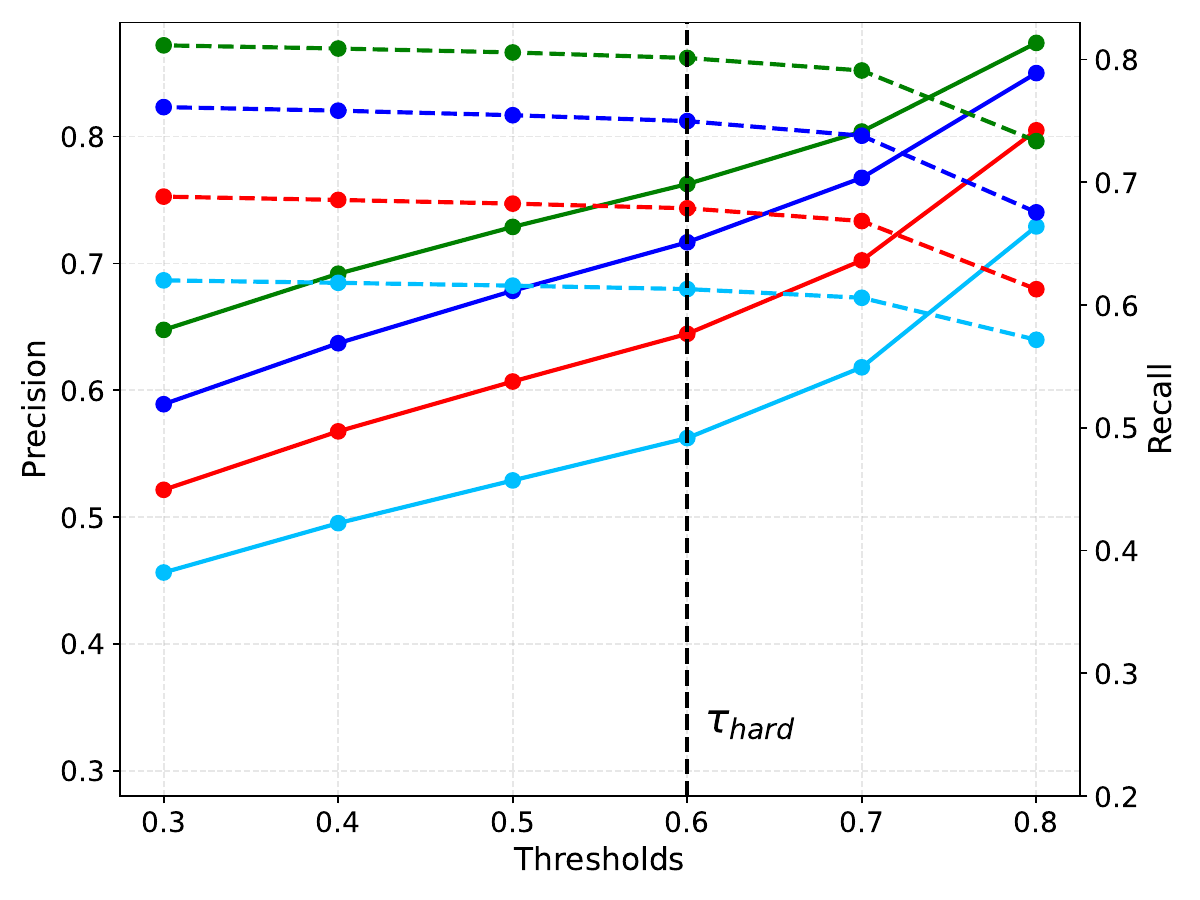}
        \vspace{-5mm}
        \caption{1Mpx WSOD Cars}
    \end{subfigure}
    \hspace{-2.5mm}
    \begin{subfigure}{0.335\linewidth}
        \includegraphics[width=1.0\linewidth]{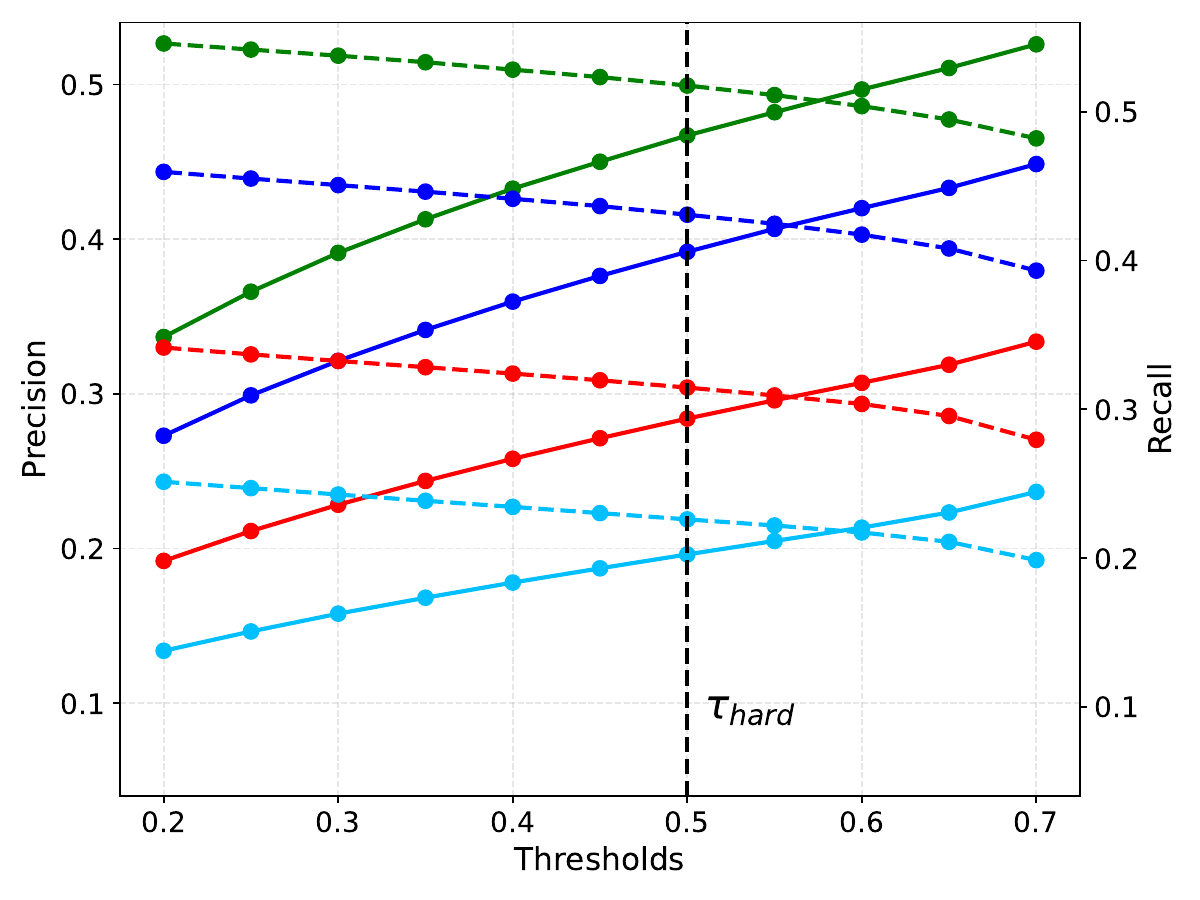}
        \vspace{-5mm}
        \caption{1Mpx WSOD Pedestrians}
    \end{subfigure}
    \hspace{-2.5mm}
    \begin{subfigure}{0.335\linewidth}
        \includegraphics[width=1.0\linewidth]{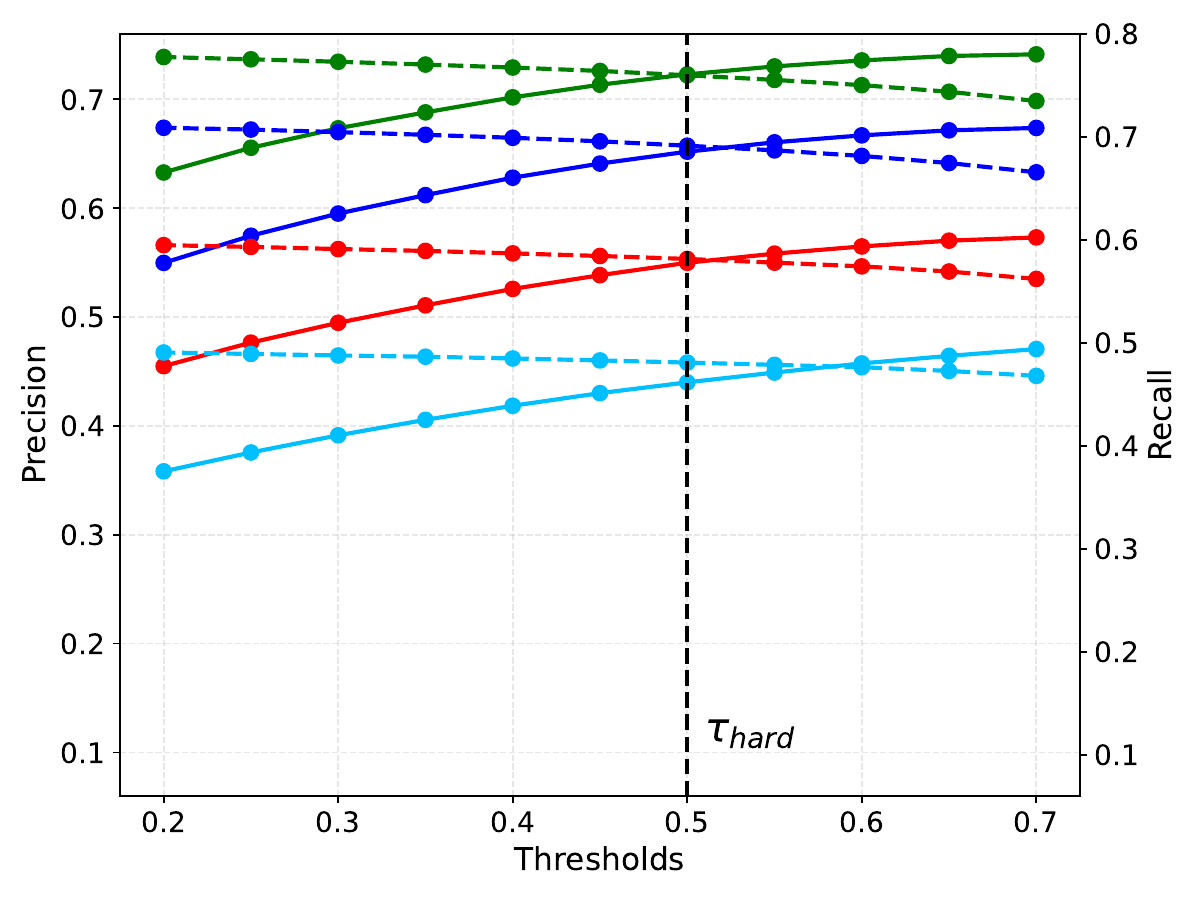}
        \vspace{-5mm}
        \caption{1Mpx WSOD \Cycs}
    \end{subfigure}
    \\ \vspace{1mm}
    \begin{subfigure}{0.335\linewidth}
        \includegraphics[width=1.0\linewidth]{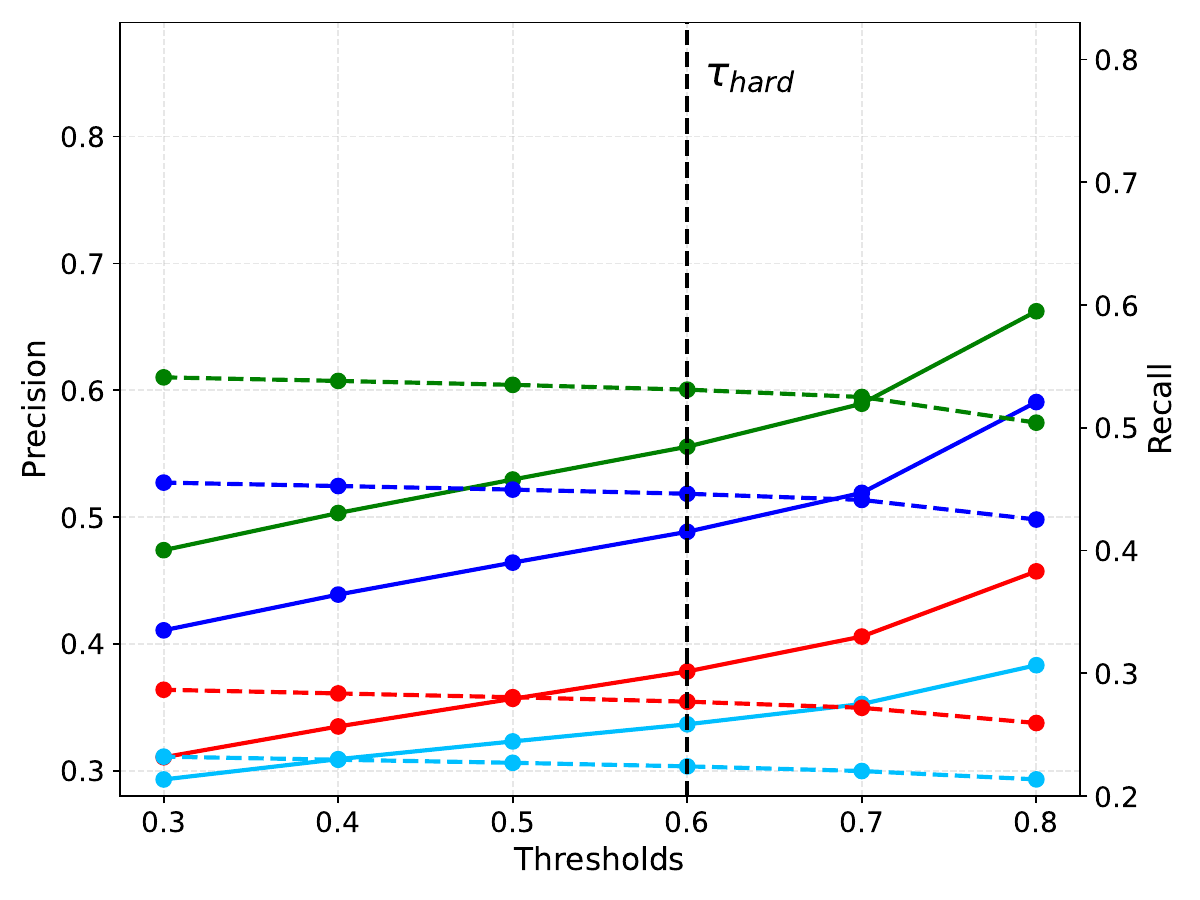}
        \vspace{-5mm}
        \caption{1Mpx SSOD Cars}
    \end{subfigure}
    \hspace{-2.5mm}
    \begin{subfigure}{0.335\linewidth}
        \includegraphics[width=1.0\linewidth]{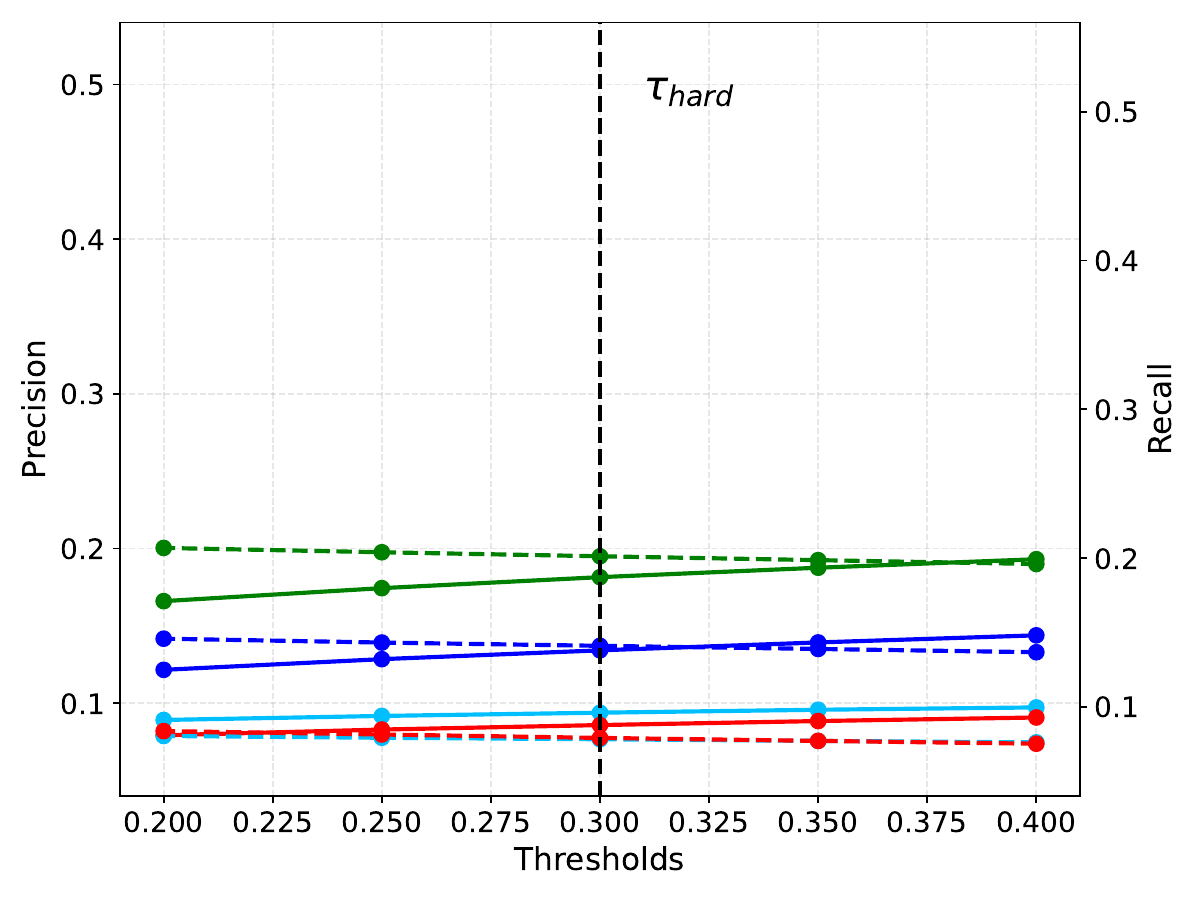}
        \vspace{-5mm}
        \caption{1Mpx SSOD Pedestrians}
    \end{subfigure}
    \hspace{-2.5mm}
    \begin{subfigure}{0.335\linewidth}
        \includegraphics[width=1.0\linewidth]{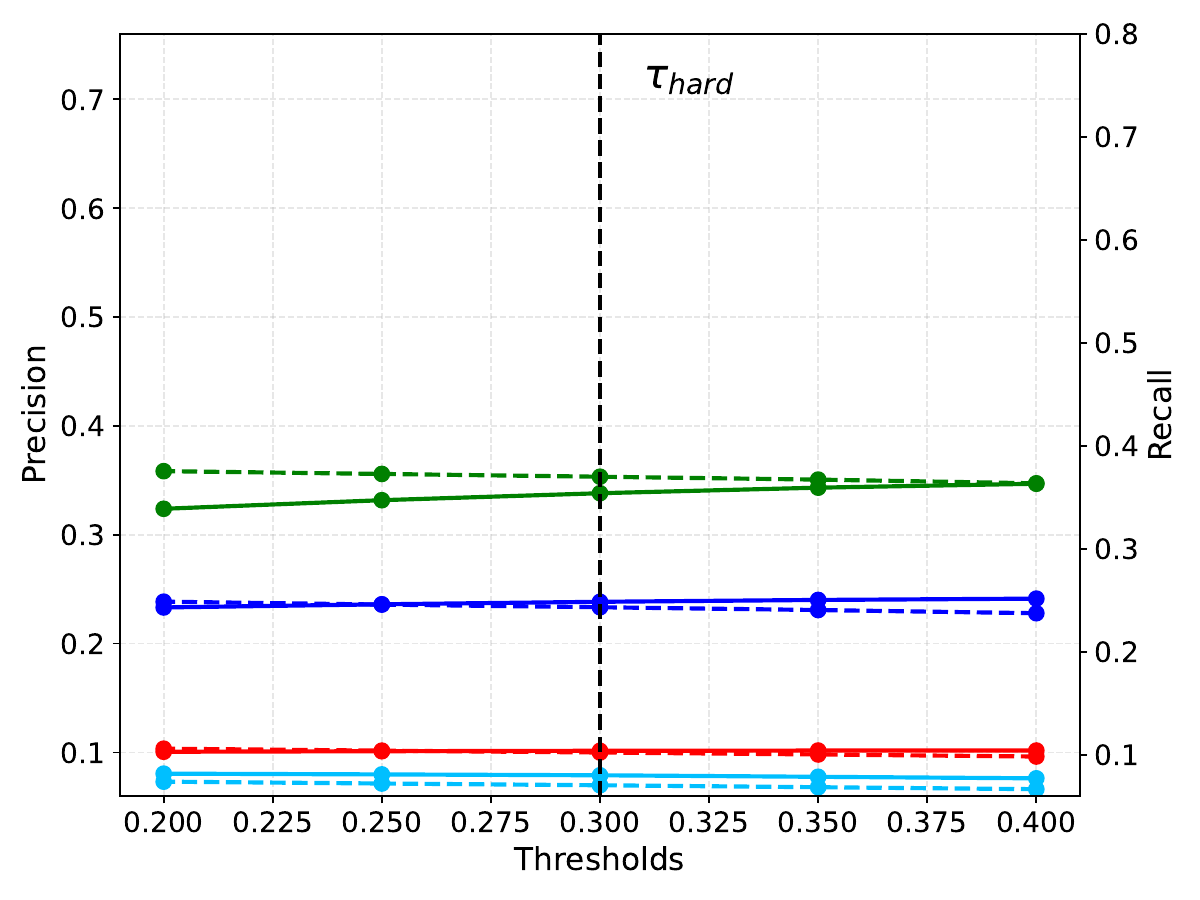}
        \vspace{-5mm}
        \caption{1Mpx SSOD \Cycs}
    \end{subfigure}
    \vspace{-2.5mm}
    \caption{
        We plot the precision and recall of pseudo labels generated under different settings.
        In each figure, solid lines represent precision and dotted lines represent recall.
        Four labeling ratios $1\%, 2\%, 5\%, 10\%$ are selected. % whose values are in ascending order.
        The black dotted line is the threshold for label filtering.
        We fix the Y-axis value range within each ground \{(a), (b), (c)\}, \{(d), (e), (f)\}, \{(g), (j)\}, \{(h), (k)\}, \{(i), (l)\} for easy comparisons.
    }
    \label{app-fig:pr-curves}
    \vspace{\figmargin}
    \vspace{2mm}
\end{figure*}

\section{Detailed Analysis of Pseudo Label Quality}\label{app:pseudo-label-pr}

\cref{app-fig:pr-curves} shows the precision and recall of pseudo labels under different settings and thresholds.
They are computed by evaluating pseudo labels against the ground-truth labels at annotated but skipped frames.
If a predicted box has an IoU higher than $0.75$ with a ground-truth box, we treat it as a positive detection.
We make the following observations:

\heading{More pre-training labels lead to better quality.}
In all settings, models pre-trained with more labels produce pseudo labels with clearly higher precision and recall.

\heading{Cars are much easier to detect than other categories.}
Comparing cars, pedestrians, and \cycs, it is clear that cars have a much better label quality in all settings.
This is because cars are larger and there are more bounding box annotations of cars than other objects.
On 1Mpx, \cycs are slightly easier to detect than pedestrians.
Future work can study how to address the class-imbalance issue and improve detections on hard examples.

% visualizing some pseudo labels
\begin{figure*}[t]
    \vspace{\pagetopmargin}
    \vspace{-2mm}
    \centering
    \includegraphics[width=0.99\linewidth]{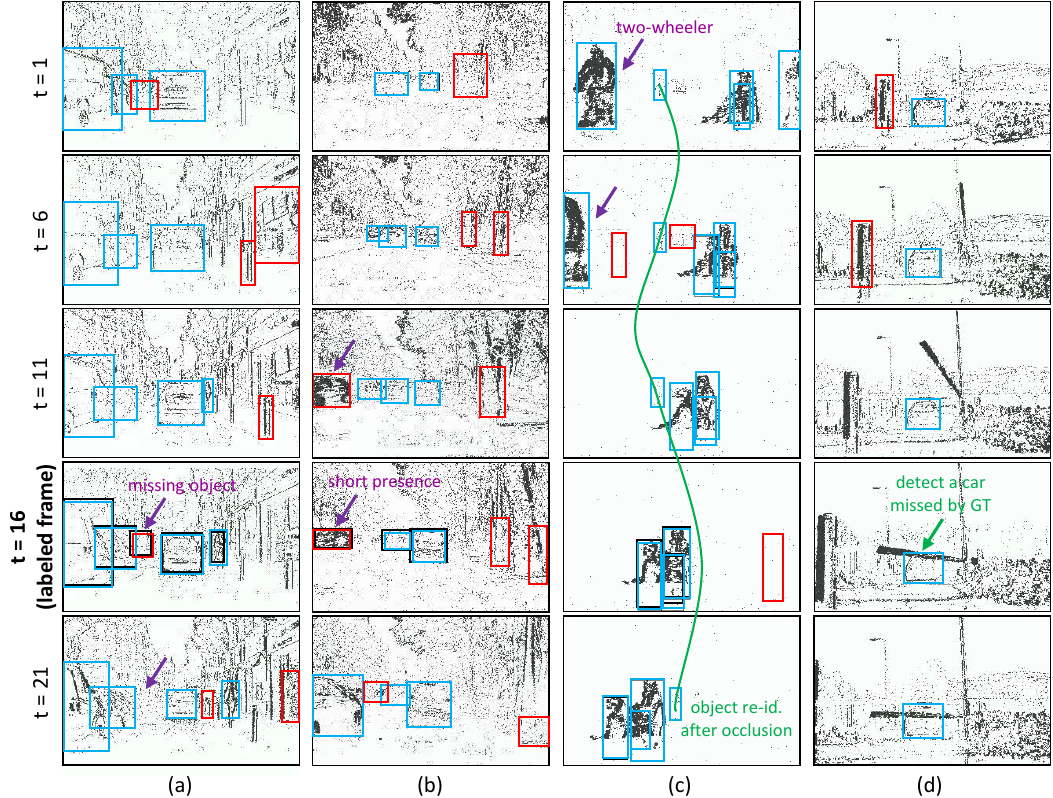}
    % \vspace{-1mm}
    \vspace{\figcapmargin}
    \caption{
        We visualize some pseudo labels on Gen1 that are generated by an RVT-S after one round of self-training.
        {\color{cyan} Blue} boxes are pseudo labels kept for model training while {\color{red} red} boxes are those removed by tracking-based post-processing.
        Black boxes at $t = 16$ are ground-truth annotations.
        The $t$ here denotes timesteps of the event frame representation instead of seconds in the real world.
        {\color{violet} Purple} arrows highlight some failure cases of our method while {\color{Green} green} arrows highlight some desired behaviors.
    }
    \label{app-fig:pse-label-vis}
    % \vspace{\figmargin}
\end{figure*}

\heading{Self-training improves pseudo label quality, but may degrade precision.}
Comparing \cref{app-fig:pr-curves} (a) and (b), (d) and (e), we can see that one round of self-training greatly improves the recall (dotted lines).
However, the precision (solid lines) drops if we use a small $\tau_\text{hard}$.
This is because the model learns to discover more objects after self-training, but is also over-confident in its predictions.
Therefore, fewer false positives are removed in the filtering process.
One solution is to increase the threshold $\tau_\text{hard}$ over the number of self-training rounds, as done in \cite{CutLer}.
We tried this in our preliminary experiments but did not observe a clear improvement.

\heading{Weakly-supervised learning (WSOD) leads to better results than semi-supervised learning (SSOD).}
Comparing the WSOD and SSOD results in \cref{app-fig:pr-curves}, we can see that models trained in WSOD produce much higher quality pseudo labels than their SSOD counterparts.
Together with the detection mAP results presented in Sec. 4.2, this proves that sparsely labeling as many event streams as possible is better than densely labeling a few event sequences.

\heading{Gen1 \vs 1Mpx.}
Comparing \cref{app-fig:pr-curves} (a) and (g), (c) and (j), it is clear that models on Gen1 detect cars much better than on 1Mpx.
This is because 1Mpx has a higher resolution and the number of cars per frame is also larger (1Mpx: 3.8 vs Gen1: 1.9).
Interestingly, as can be seen from \cref{app-fig:pr-curves} (d) and (h), the label quality of pedestrians on Gen1 is worse than on 1Mpx.
After visualizing some results, we realize that this is because Gen1 does not provide annotations for \cycs, but the model detects lots of \cycs as pedestrians, which are regarded as false positives.
In contrast, 1Mpx does not have this issue as \cycs are also labeled which disambiguates model learning.
Indeed, the gap in precision is much higher than recall, as precision penalizes false positives.
Future work can study how to learn more discriminative features to separate object categories, \eg with class-centric contrastive loss~\cite{3DLiDARLESS}.

\section{Visualization of Pseudo Labels}\label{app:pseudo-label-vis}

We visualize some pseudo labels on Gen1 in \cref{app-fig:pse-label-vis}.

\heading{Failure case analysis.}
Tracking-based post-processing is able to eliminate temporally inconsistent boxes.
However, since we use a fixed threshold $T_{trk} = 6$ for all tracks, some objects may be incorrectly removed.
In \cref{app-fig:pse-label-vis} (a), the car highlighted by the purple arrow is a hard example as it only triggers a few events.
The model only detects it in one frame while missing it in later frames, leading to a short track length.
As a result, the correct detection at $t = 16$ is mistakenly removed.
In \cref{app-fig:pse-label-vis} (b), the cars coming from the other direction move very fast, and only stay visible for less than $T_{trk}$ timesteps.
Thus, they are also wrongly removed.
Nevertheless, since we ignore these boxes during model training instead of suppressing them as background, such errors are less harmful.
\cref{app-fig:pse-label-vis} (c) shows another failure case where a \cyc is recognized as a pedestrian as discussed in \cref{app:pseudo-label-pr}.

\heading{Successful examples.}
In \cref{app-fig:pse-label-vis} (c), we visualize the tracking trajectory of a pedestrian (the green curve).
Although the pedestrian is occluded and thus not detected at $t = 16$, our tracker is able to re-identify it at $t = 21$, thus keeping it in the pseudo labels.
\cref{app-fig:pse-label-vis} (d) shows an example where a car is not annotated in the ground-truth labels.
Our model successfully discovers it and corrects the annotation error.

\section{Discussion on Experimental Setting Naming}\label{app:discuss-wsod-ssod}

In this paper, we propose two settings under the label-efficient event-based detection task:
\textbf{(i)} weakly-supervised object detection (WSOD) where all event sequences are sparsely annotated, and \textbf{(ii)} semi-supervised object detection (SSOD) where some event sequences are densely annotated, and others are fully unlabeled.
While (ii) undoubtedly belongs to semi-supervised learning, (i) may be controversial.
In fact, the definition of weakly- and semi-supervised learning is often overlapping in the literature.
For example, the Wikipedia page\footnote{\url{https://en.wikipedia.org/wiki/Weak_supervision}} seems to give similar definitions to these two tasks: ``\textbf{Weak supervision}, also called \textbf{semi-supervised learning}, is a paradigm in machine learning..."
Previous surveys~\cite{semi-sup-survey1,semi-sup-survey2} identify a key property in semi-supervised learning: labeled and unlabeled data should be (although from the same distribution) independent of each other.
In contrast, the labeled frames in a sparsely labeled event sequence are not independent of the unlabeled frames in the same sequence.
On the other hand, another survey on weakly-supervised learning~\cite{weakly-sup-survey1} regards ``incomplete supervision where only a subset of training data are given with labels" as one type of weak supervision, which is similar to our sparse labeling setting.
These are the main reasons we term (i) weakly-supervised learning to differentiate it from semi-supervised learning.

However, we note that some works~\cite{SemiSupVidDet1,SemiSupVidDet2} learning video object detection with sparsely labeled frames call their setting semi-supervised learning.
Moreover, if we employ a feedforward detector, \ie detectors that do not leverage temporal information, setting (i) becomes closer to semi-supervised learning as labeled and unlabeled timesteps become less relevant.
Nevertheless, we believe recurrent detectors are the future trend in event-based object detection as they lead to significantly stronger performance.

\section{Societal Impact}\label{app:ethics}

This paper proposes a framework to learn better event-based object detectors with limited labeled data.
Object detection is a core task in computer vision that is used across a wide variety of applications including healthcare, entertainment, communication, mobility, and defense.
While only a subset of scenarios in this application can benefit from event-camera data, it is still difficult to predict the overall impact of the technology.
Moreover, event-based detectors may introduce biases that are different from those encountered in classical cameras and better understanding such biases is an open research problem.
While we do not see any immediate risks of human rights or security violations introduced by our work, future work building upon it will carefully need to investigate implications on its particular application area.

\end{document}